%% file: main.tex
\documentclass{article}

\usepackage{microtype}
\usepackage{graphicx}
\usepackage{subfigure}
\usepackage{booktabs} 
\usepackage{amsmath}

\usepackage{hyperref}

\newcommand{\jenny}[1]{{\textcolor{blue}{[Jenny: #1]}}}
\newcommand{\ili}[1]{{\textcolor{cyan}{[Ili: #1]}}}
\newcommand{\lau}[1]{\textcolor{magenta}{Laura: #1}}

\usepackage[accepted]{icml2021}

\icmltitlerunning{Learning Intra-Batch Connections for Deep Metric Learning}

\begin{document}

\twocolumn[
\icmltitle{Learning Intra-Batch Connections for Deep Metric Learning}




\begin{icmlauthorlist}
\icmlauthor{Jenny Seidenschwarz}{tum}
\icmlauthor{Ismail Elezi}{tum}
\icmlauthor{Laura Leal-Taixé}{tum}
\end{icmlauthorlist}

\icmlaffiliation{tum}{Department of Computer Science, Technical University of Munich, Munich, Germany}

\icmlcorrespondingauthor{Jenny Seidenschwarz}{j.seidenschwarz@tum.de}

\icmlkeywords{Deep Metric Learning, ICML}

\vskip 0.3in
]



\printAffiliationsAndNotice{}  

\begin{abstract}
    %
    The goal of metric learning is to learn a function that maps samples to a lower-dimensional space where similar samples lie closer than dissimilar ones. Particularly, deep metric learning utilizes neural networks to learn such a mapping. 
    %
    Most approaches rely on losses that only take the relations between pairs or triplets of samples into account, which either belong to the same class or two different classes.  
    However, these methods do not explore the embedding space in its entirety. 
    To this end, we propose an approach based on message  passing networks that takes all the relations in a mini-batch into account. We refine embedding vectors by exchanging messages among all samples in a given batch allowing the training process to be aware of its overall structure.
    Since not all samples are equally important to predict a decision boundary, we use an attention mechanism 
    during message passing to allow samples to weigh the importance of each neighbor accordingly.
    We achieve state-of-the-art results on clustering and image retrieval on the CUB-200-2011, Cars196, Stanford Online Products, and In-Shop Clothes datasets. To facilitate further research, we make available the code and the models at \url{https://github.com/dvl-tum/intra_batch}.

\end{abstract}

\input{tex_files/intro}

\input{tex_files/related_work}
\input{tex_files/methodology}

\input{tex_files/experiments}

\input{tex_files/conclusion}
\input{tex_files/appendix}

\clearpage
\bibliographystyle{icml2021}
\bibliography{bibliography}

\end{document}

%% file: tex_files/intro.tex
\section{Introduction}
Metric learning is a widely popular technique that constructs task-specific distance metrics by learning the similarity or dissimilarity between samples.
It is often used for object retrieval and clustering by training a deep neural network to learn a mapping function from the original samples into a new, more compact, embedding space. %
In that embedding space, samples coming from the same class should be closer than samples coming from different classes.

\begin{figure*}[hbt!]
\begin{center}
\centerline{\includegraphics[width=0.95\linewidth]{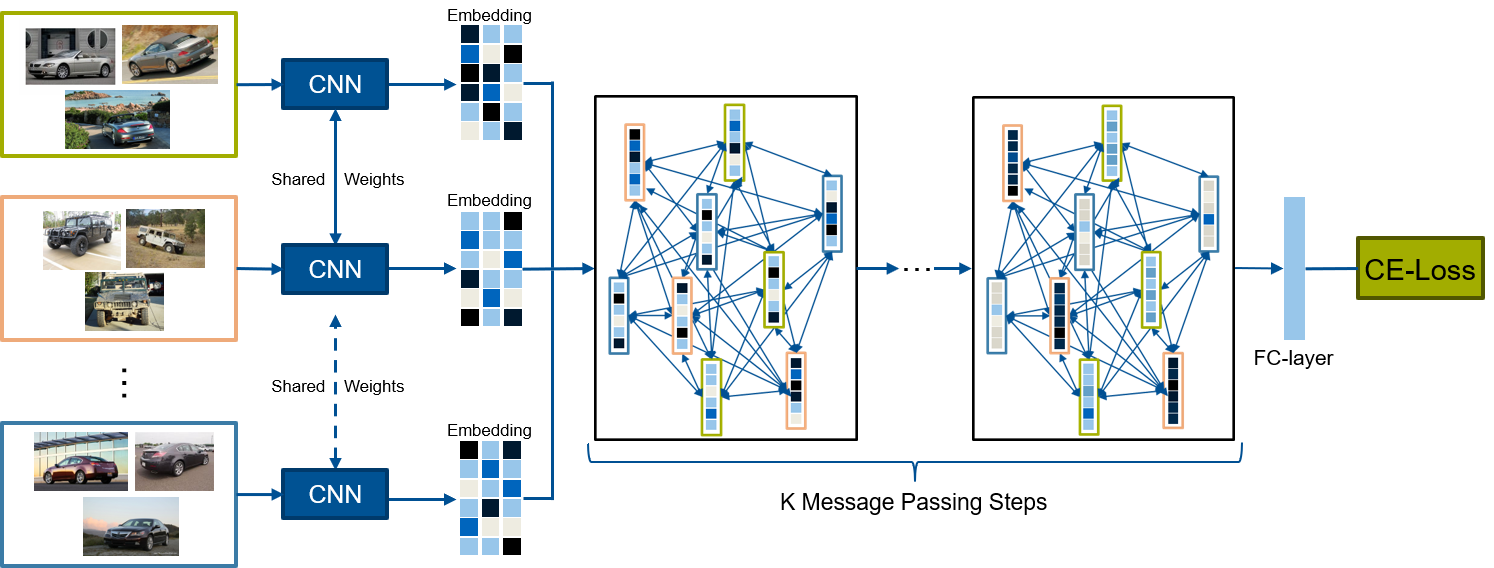}}
   \caption{Overview of our proposed approach. Given a mini-batch consisting of $N$ classes, each of them having $P$ images, we initialize the embedding vectors using a backbone CNN. We then construct a fully connected graph that refines their initial embeddings by performing $K$ message-passing steps. After each step, the embeddings of the images coming from the same class become more similar to each other and more dissimilar to the embeddings coming from images that belong to different classes. Finally, we apply Cross-Entropy loss and we backpropagate the gradients to update the network.}
\label{fig:message}
\end{center}
\end{figure*}

To learn the mapping function, current approaches utilize siamese networks \cite{bromley1994signature}, typically trained using loss functions that measure distances between pairs of samples of the same class (positive) or different classes (negative). 
Contrastive loss \cite{bromley1994signature} minimizes the distance of the feature embeddings for a positive pair, and maximizes their distance otherwise.
Triplet loss \cite{DBLP:conf/nips/SchultzJ03,DBLP:journals/jmlr/WeinbergerS09} takes a triplet of images and pushes the embedding distance between an anchor and a positive sample to be smaller than the distance between the same anchor and a negative sample by a given margin.
%
While the number of possible image pairs and triplets in a dataset of size $n$ is $\mathcal{O}(n^2)$ and $\mathcal{O}(n^3)$, respectively, the vast majority of these pairs (or triplets) are not informative and do not contribute to the loss. 
This leads to slow convergence and possible overfitting when the pairs (triplets) are not appropriately sampled. Perhaps more worryingly, because these losses are focused on pairs (triplets), they are unable to consider the global structure of the dataset resulting in lower clustering and retrieval performance. 
To compensate for these drawbacks, several works resort to training tricks like intelligent sampling \cite{DBLP:conf/eccv/GeHDS18,DBLP:conf/iccv/ManmathaWSK17}, multi-task learning \cite{DBLP:conf/cvpr/ZhangZLZ16}, or hard-negative mining \cite{DBLP:conf/cvpr/SchroffKP15, DBLP:journals/corr/abs-2007-12749}. 
Recently, researchers started exploring the global structure of the embedding space by utilizing rank-based \cite{DBLP:conf/cvpr/Cakir0XKS19,DBLP:conf/cvpr/0003CBS18,DBLP:journals/corr/abs-1906-07589} or contextual classification loss functions \cite{DBLP:conf/cvpr/Cakir0XKS19,DBLP:conf/eccv/GrLoss,DBLP:conf/cvpr/0003CBS18,DBLP:journals/corr/abs-1906-07589,DBLP:conf/nips/Sohn16,DBLP:conf/cvpr/SongXJS16,DBLP:conf/aaai/ZhengJSZWH19}. The Group Loss \cite{DBLP:conf/eccv/GrLoss} explicitly considers the global structure of a mini-batch and refines class membership scores based on feature similarity. However, the global structure is captured using a handcrafted rule instead of learning, hence its refinement procedure cannot be adapted depending on the samples in the mini-batch.

\subsection{Contributions}

In this work, we propose a fully learnable module that takes the global structure into account by refining the embedding feature vector of each sample based on \textit{all} intra-batch relations.
To do so, we utilize message passing networks (MPNs) \cite{DBLP:conf/icml/GilmerSRVD17}. MPNs allow the samples in a mini-batch to communicate with each other, and to refine their feature representation based on the information taken from their neighbors. 
More precisely, we use a convolutional neural network (CNN) to generate feature embeddings. 
We then construct a fully connected graph where each node is represented by the embedding of its corresponding sample. 
In this graph, a series of message passing steps are performed to update the node embeddings. 
Not all samples are equally important to predict decision boundaries, hence, we allow each sample to weigh the importance of neighboring samples by using a dot-product self-attention mechanism to compute aggregation weights for the message passing steps.

To draw a parallelism with the triplet loss, our MPN formulation would allow samples to choose their own triplets which are best to make a prediction on the decision boundary. Unlike the triplet loss though, we are not limited to triplets, as each sample can choose to attend over all other samples in the mini-batch.
%
By training the CNN and MPN in an end-to-end manner, we can directly use our CNN backbone embeddings during inference to perform image retrieval and clustering.
While this reaches state-of-the-art results without adding any computational overhead, we also show how to further boost the performance by using the trained MPN at test time, constructing the batches based on k-reciprocal nearest neighbor sampling \cite{DBLP:conf/cvpr/ZhongZCL17}.

Our \textbf{contribution} in this work is three-fold: 
\begin{itemize}
    \item We propose an approach for deep metric learning that computes sample embeddings by taking into account all intra-batch relations. By leveraging message passing networks, our method can be trained end-to-end.
    \item We perform a comprehensive robustness analysis showing the stability of our module with respect to the choice of hyperparameters.
    \item We present state-of-the-art results on CUB-200-2011 \cite{WahCUB_200_2011}, Cars196 \cite{KrauseStarkDengFei-Fei_3DRR2013}, Stanford online Products \cite{DBLP:conf/cvpr/SongXJS16} and In-Shop Clothes \cite{DBLP:conf/cvpr/LiuLQWT16} datasets. 
\end{itemize}

%% file: tex_files/related_work.tex
\section{Related Work}

\textbf{Metric Learning Losses.} Siamese neural networks were first proposed for representation learning in~\cite{bromley1994signature}. The main idea is to use a CNN to extract a feature representation from an image and using that representation, or embedding, to compare it to other images.
In \cite{DBLP:conf/cvpr/ChopraHL05}, the contrastive loss was introduced to train such a network for face verification. The loss minimizes the distance between the embeddings of image pairs coming from the same class and maximizes the distance between image pairs coming from different classes.
In parallel, researchers working on convex optimization developed the triplet loss \cite{DBLP:conf/nips/SchultzJ03,DBLP:journals/jmlr/WeinbergerS09} which was later combined with the expressive power of CNNs, further improving the solutions on face verification \cite{DBLP:conf/cvpr/SchroffKP15}. 
Triplet loss extends contrastive loss by using a triplet of samples consisting of an anchor, a positive, and a negative sample, where the loss is defined to make the distance between the anchor and the positive smaller than the distance between the anchor and the negative, up to a margin. 
The concept was later generalized to N-Pair loss \cite{DBLP:conf/nips/Sohn16}, where an anchor and a positive sample are compared to $N-1$ negative samples at the same time. In recent years, different approaches based on optimizing other qualities than the distance, such as clustering \cite{DBLP:conf/icml/LawUZ17,DBLP:journals/corr/abs-1110-2515} or angular distance \cite{DBLP:conf/iccv/WangZWLL17}, have shown to reach good results. 

\textbf{Sampling and Ensembles.} Since computing the loss of all possible triplets is computationally infeasible even for moderately-sized datasets and, furthermore, based on the knowledge that the majority of them are not informative \cite{DBLP:conf/cvpr/SchroffKP15}, more researchers have given attention to intelligent \textit{sampling}. 
The work of \cite{DBLP:conf/iccv/ManmathaWSK17} showed conclusive evidence that the design of smart sampling strategies is as important as the design of efficient loss functions.
In \cite{DBLP:conf/eccv/GeHDS18}, the authors propose a hierarchical version of triplet loss that embeds the sampling during the training process. More recent techniques continue this line of research by developing new sampling strategies \cite{DDBLP:conf/cvpr/Duan2019, DBLP:journals/corr/abs-2007-12749,DBLP:conf/wacv/XuanSP20} while others introduce new loss functions \cite{DDBLP:conf/cvpr/Wand2019,DDBLP:conf/cvpr/Xu2019}. 
%
In parallel, other researchers investigated the usage of ensembles for deep metric learning, unsurprisingly finding out that ensembles outperform single networks trained on the same loss \cite{DBLP:conf/eccv/KimGCLK18,DBLP:conf/iccv/OpitzWPB17,DDBLP:conf/cvpr/Sanakoyeu2019,DBLP:conf/eccv/XuanSP18,DBLP:conf/iccv/YuanYZ17}.

\textbf{Global Metric Learning Losses.} Most of the mentioned losses do not consider the global structure of the mini-batch. 
The work of \cite{DBLP:conf/iccv/Movshovitz-Attias17} proposes to optimize the triplet loss on a space of triplets different from the one of the original samples, consisting of an anchor data point and similar and dissimilar learned proxy data points. 
These proxies approximate the original data points so that a triplet loss over the proxies is a tight upper bound of the loss over the original samples. 
The introduction of proxies adds additional contextual knowledge that shows to significantly improve triplet loss. The results of this approach were significantly improved by using training tricks \cite{DBLP:journals/corr/abs-2004-01113} or generalizing the concept of proxy triplets to multiple proxy anchors \cite{DBLP:conf/cvpr/KimKCK20, DBLP:journals/corr/abs-2010-13636}. 
In \cite{DBLP:conf/cvpr/DuanZLL018} the authors generate negative samples in an adversarial manner, while in \cite{DBLP:conf/eccv/LinDDLZ18} a deep variational metric learning framework was proposed to explicitly model the intra-class variance and disentangle the intra-class invariance.
In the work of \cite{DBLP:conf/cvpr/WangHKHGR19}, a non-proxy contextual loss function was developed. The authors propose a loss function based on a ranking distance that considers all the samples in the mini-batch 
%

\textbf{Classification Losses for Metric Learning.} A recent line of work~\cite{DBLP:journals/corr/abs-1811-12649, DBLP:conf/aaai/ZhengJSZWH19} is showing that a carefully designed classification loss function can rival, if not outperform, triplet-based functions in metric learning. This has already been shown for multiple tasks such as hashing (binary-embedding) \cite{DBLP:conf/cvpr/0003CBS18}, landmark detection \cite{DBLP:conf/cvpr/0003LS18, DBLP:journals/corr/abs-1906-07589}, few-shot learning \cite{DBLP:conf/cvpr/Cakir0XKS19}, and person re-identification \cite{DBLP:journals/corr/abs-1904-11397,DBLP:conf/cvpr/ZhaoXC19}. In metric learning, SoftTriple loss \cite{DBLP:journals/corr/abs-1909-05235} develops a classification loss where each class is represented by $K$ centers. In the same classification spirit, the Group Loss \cite{DBLP:conf/eccv/GrLoss} replaces the softmax function with a contextual module that considers all the samples in the mini-batch at the same time.  

\textbf{Message Passing Networks.}
Recent works on message passing networks \cite{DBLP:conf/icml/GilmerSRVD17} and graph neural networks \cite{DBLP:journals/corr/abs-1806-01261,DBLP:conf/iclr/KipfW17} have been successfully applied to problems such as human action recognition \cite{DBLP:conf/eccv/GuoCHSYF18}, visual question answering \cite{DBLP:conf/nips/NarasimhanLS18} or tracking \cite{DBLP:conf/cvpr/BrasoL20}. 
Given a graph with some initial
features for nodes and edges, the main idea behind these models is to embed nodes and edges into representations that take into account not only the node’s own
features but also those of its neighbors in the graph, as well as the graphs overall topology. 
The attention-based Transformers \cite{DBLP:conf/nips/VaswaniSPUJGKP17,DBLP:conf/icml/XuBKCCSZB15}, which can be seen as message passing networks, have revolutionized the field of natural language processing, and within the computer vision, have shown impressive results in object detection \cite{DBLP:journals/corr/abs-2005-12872}.

Closely related to message passing networks, \cite{DBLP:conf/eccv/GrLoss} considered contextual information for metric learning based on the similarity (dissimilarity) between samples coming from the same class (respectively from different classes). However, they use a handcrafted rule as part of their loss function that only considers the label preferences \cite{DBLP:conf/icpr/EleziTVP18}. In contrast, based on message passing networks, we develop a novel learnable model, where each sample uses learned attention scores to choose the importance of its neighbors, and based on this information, refines its own feature representation.

%% file: tex_files/methodology.tex
\section{Methodology}
The goal of the message passing steps is to exchange information between all samples in the mini-batch and to refine the feature embeddings accordingly. 
Note that this approach is very different from label-propagation methods as used in \cite{DBLP:conf/eccv/GrLoss}, where samples exchange information only on their label preferences, information which only implicitly affects the choice of their final feature vectors.

In our proposed method, each sample exchanges messages with all the other samples in the mini-batch, regardless of whether the samples belong to the same class or not. In this way, \textit{our method considers both the intra-class and inter-class relations between all samples in the mini-batch}, allowing our network to receive information about the overall structure of the mini-batch.
%
%
%
%
We can use cross-entropy loss to train our network since the information of the mini-batch is already contained in the refined individual feature embeddings. 

\subsection{Overview}
\label{subsec:overview}

\begin{figure*}
\centerline{\includegraphics[width=0.95\linewidth]{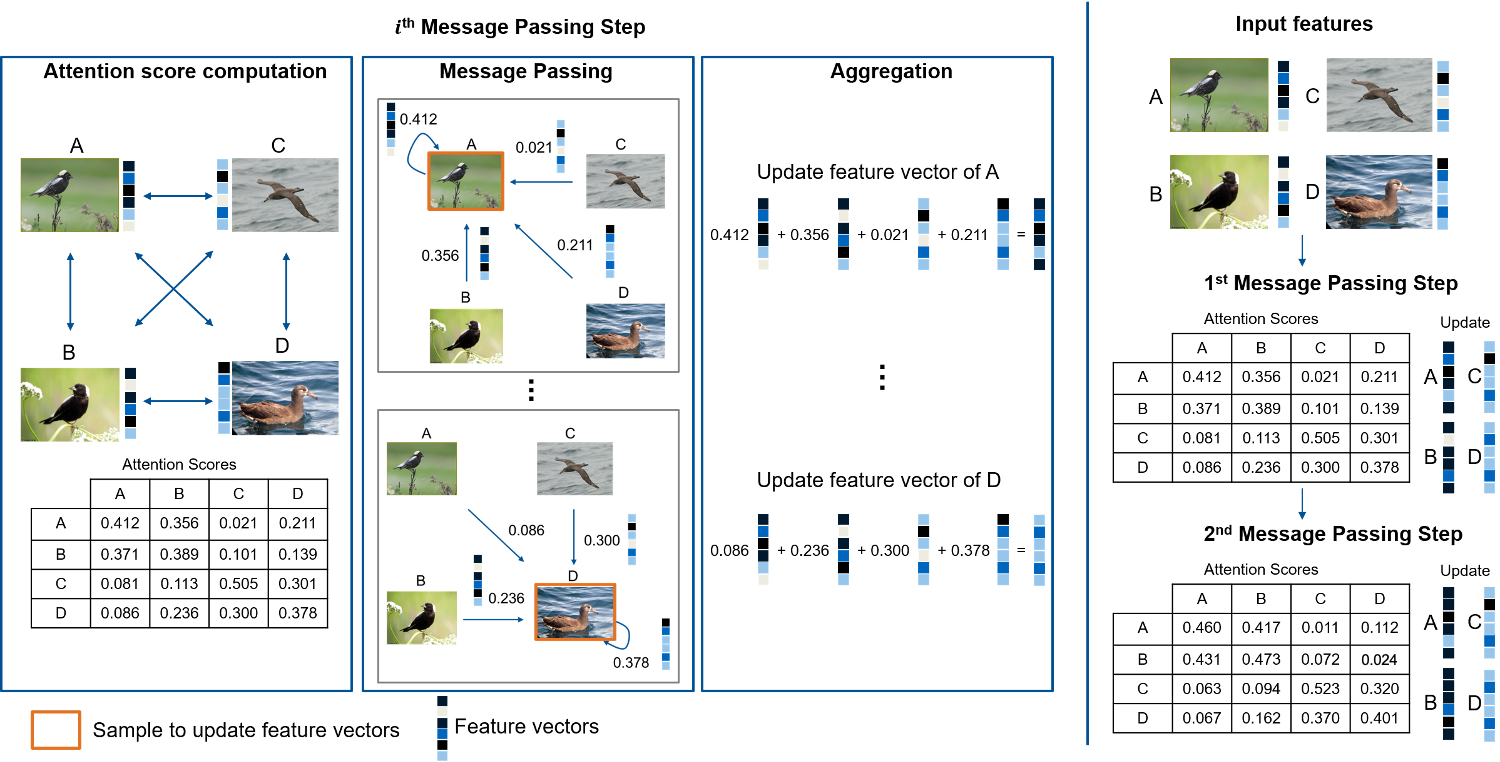}}
   \caption{Left: To update the feature vectors in a message-passing step we first construct a fully connected graph and compute attention scores between all samples in a batch. We then pass messages between nodes and weigh them with the corresponding attention scores. During the aggregation step, we sum the weighted messages to get updated node features. 
   Right: Visualization of development of attention scores and feature vectors over two steps of message passing steps showing that feature vectors, as well as attention scores between samples from the same class, get more and more similar.}
   \vspace{+0.3cm}
\label{fig:mpn2}
\end{figure*}


In Figure~\ref{fig:message}, we show an overview of our proposed approach. We compute feature vectors for each sample as follows:

\begin{enumerate}
    \item Generate initial embedding feature vectors using a CNN and construct a fully connected graph, where each node represents a sample in the mini-batch.
    \item Perform message-passing between nodes to refine the initial embedding feature vectors by utilizing dot-product self-attention.
    \item Perform classification and optimize both the MPN and the backbone CNN in an end-to-end fashion using cross-entropy loss on the refined node feature vectors. 
\end{enumerate}
\vspace{-0.3cm}
\subsection{Feature Initialization and Graph Construction}
\label{subsec:init}

The global structure of the embedding space is modeled by a graph $\mathcal{G} = (V, E)$, where $V$ represents the nodes, i.e., all images in the training dataset, and $E$ the edges connecting them. An edge represents the importance of one image to the other, expressed, for example, by their similarity.
%
During training, we would ideally take the graph of the whole dataset into account, but this is computationally infeasible. Therefore, we construct mini-batches consisting of $n$ randomly sampled classes with $p$ randomly chosen samples per class. Each sample in the mini-batch is regarded as a node in a mini-batch graph $\mathcal{G_B} = (V_B, E_B)$.
Unlike CNNs that perform well on data with an underlying grid-like or Euclidean structure \cite{GeoDeepL}, graphs have a non-euclidean structure. Thus, to fully explore the graph-like structure, we model the mini-batch relations using MPNs. 

More precisely, we use a backbone CNN to compute the initial embeddings $\boldsymbol{f} \in \mathcal{R}^{d}$ for all samples in a mini-batch, where ${d}$ is their embedding dimension. To leverage all relations in the batch, we utilize a fully connected graph, where every node with initial node features $\boldsymbol{h}_i^0= \boldsymbol{f}$ is connected to all the other nodes in the graph (see Figure~\ref{fig:mpn2} in the upper left corner). 
\vspace{-0.3cm}
\subsection{Message Passing Network}
\label{subsec:MPN}
In order to refine the initial feature vectors based on the contextual information of the mini-batch, we use message passing to exchange information between single nodes, i.e., between samples of the mini-batch. 
To this end, we utilize MPNs with graph attention \cite{DBLP:conf/iclr/VelickovicCCRLB18} for deep metric learning. It should be noted that the following formulation is equivalent to the Transformers architecture \cite{DBLP:conf/nips/VaswaniSPUJGKP17}, which can be seen as a fully connected graph attention network~\cite{DBLP:conf/iclr/VelickovicCCRLB18}.

\noindent\textbf{Passing Messages.} We apply $L$ message passing steps successively. In each step, we pass messages between all samples in a batch and obtain updated features $\boldsymbol{h}_i^{l+1}$ of node $i$ at message passing step $l+1$ by aggregating the features $\boldsymbol{h}_j^l$ of all neighbouring nodes $ j \in N_i$ at message passing step $l$:
\begin{equation}
    \boldsymbol{h}_i^{l+1} = \sum_{j \in N_i} \boldsymbol{W}^l\boldsymbol{h}_j^l
\end{equation}

where $\boldsymbol{W}^l$ is the corresponding weight matrix of message passing step $l$. As we construct a fully connected graph, the neighboring nodes $N_i$ consist of all nodes in the given batch, thus each feature representation of an image is affected by all the other images in the mini-batch.

\noindent\textbf{Attention Weights on the Messages.} 
Not all samples of a mini-batch are equally informative to predict the decision boundaries between classes. Hence, we add an attention score $\alpha$ to every message passing step (see Figure~\ref{fig:mpn2} on Message Passing) to allow each sample to weigh the importance of the other samples in the mini-batch:
\begin{equation}
    \boldsymbol{h}_i^{l+1} = \sum_{j \in N_i} \alpha_{ij}^l \boldsymbol{W}^l\boldsymbol{h}_j^l
\end{equation}

where $\alpha_{ij}$ is the attention score between node $i$ and node $j$. 
We utilize dot-product self-attention to compute the attention scores, leading to $\alpha_{ij}$ at step $l$ defined as:
\begin{equation}
\label{equ:alpha}
    \alpha_{ij}^l = \frac{\boldsymbol{W}_q^l \boldsymbol{h}_i^l (\boldsymbol{W}_k^l \boldsymbol{h}_j^l)^T}{\sqrt{d}}
\end{equation}

where $\boldsymbol{W}_q^l$ is the weight matrix corresponding to the receiving node and $\boldsymbol{W}_k^l$ is the weight matrix corresponding to the sending node on message passing step $l$. Furthermore, we apply the softmax function to all in-going attention scores (edges) of a given node $i$. 
To allow the MPN to learn a diverse set of attention scores, we apply $M$ dot product self-attention heads in every message passing step and concatenate their results. To this end, instead of using single weight matrices $\boldsymbol{W}_q^l$, $\boldsymbol{W}_k^l$ and $\boldsymbol{W}^l$, we now use different weight matrices $\boldsymbol{W}_q^{l, m} \in \mathcal{R}^{\frac{d}{M} \times d}$, $\boldsymbol{W}_k^{l, m} \in \mathcal{R}^{\frac{d}{M} \times d}$ and $\boldsymbol{W}^{l, m} \in \mathcal{R}^{\frac{d}{M} \times d}$ for each attention head: 
\begin{equation}
\label{equ:multihead}
    \boldsymbol{h}_i^{l+1}  = cat(\sum_{j \in N_i} \alpha_{ij}^{l,1} \boldsymbol{W}^{l,1}\boldsymbol{h}_j^l, ..., \sum_{j \in N_i} \alpha_{ij}^{l,M} \boldsymbol{W}^{l,M}\boldsymbol{h}_j^l) \\
\end{equation}
where $cat$ represents the concatenation.

Note, by using the attention-head specific weight matrices, we reduce the dimension of all embeddings $\boldsymbol{h}_j^l$ by $\frac{1}{M}$ so that when we concatenate the embeddings generated by all attention heads the resulting embedding $\boldsymbol{h}_i^{l+1}$ has the same dimension as the input embedding $\boldsymbol{h}_i^{l}$.

\noindent\textbf{Adding Skip Connections.} We add a skip connection around the attention block \cite{7780459} and apply layer normalization \cite{DBLP:journals/corr/BaKH16} given by:
\begin{equation}
\label{equ:norm1res1}
    f(\boldsymbol{h}_i^{l+1}) = LayerNorm(\boldsymbol{h}_i^{l+1} + \boldsymbol{h}_i^{l})
\end{equation}

where $\boldsymbol{h}_i^{l+1}$ is the outcome of Equation~\ref{equ:multihead}. 
%
%
We then apply two fully connected layers, followed by another skip connection \cite{7780459} and a layer normalization \cite{DBLP:journals/corr/BaKH16}: 
\begin{equation}
\label{equ:norm2res2}
    g(\boldsymbol{h}_i^{l+1}) = LayerNorm(FF(f(\boldsymbol{h}_i^{l+1})) + f(\boldsymbol{h}_i^{l+1}))
\end{equation}

where $FF$ represents the two linear layers. Finally, we pass the outcome of Equation~\ref{equ:norm2res2} to the next message passing step. 
For illustrative purposes, in Figure~\ref{fig:mpn2}, we show how the attention scores and the feature vectors evolve over the message passing steps. 
As can be seen, the feature vectors of samples of the same class become more and more similar. 
%
Similar to \cite{DBLP:conf/iclr/VelickovicCCRLB18}, we indirectly address oversmoothing by applying node-wise attention scores $\alpha_{i,j }$ (Equation~\ref{equ:alpha}) during the feature aggregation step \cite{DBLP:conf/nips/Min2020}. 

\vspace{-0.3cm}
\subsection{Optimization}

\label{subsec:optimization}
We apply a fully connected layer on the refined features after the last message passing step and then use cross-entropy loss.
Even if cross-entropy loss itself does not take into account the relations between different samples, this information is already present in the refined embeddings, thanks to the message passing steps. 
As the MPN takes its initial feature vectors from the backbone CNN, we add an auxiliary cross-entropy loss to the backbone CNN, to ensure a sufficiently discriminative initialization. This loss is also needed since at test time we do not use the MPN, as described below. 
Both loss functions utilize label smoothing 
and low temperature scaling \cite{DBLP:journals/corr/abs-2004-01113,DBLP:journals/corr/abs-1811-12649} to ensure generalized, but discriminative, decision boundaries.
\vspace{-0.3cm}
\subsection{Inference}
One disadvantage of using the MPN during inference is that in order to generate an embedding vector for a sample, we need to create a batch of samples to perform message passing as we do during the training. 
However, using the MPN during inference would be unfair to other methods that directly perform retrieval on the CNN embedding since we would be adding parameters, hence, expressive power, to the model. 
Therefore, we perform all experiments by directly using the embedding feature vectors of the backbone CNN unless stated differently. The intuition is that when optimizing the CNN and MPN together in an end-to-end fashion, the CNN features will have also improved with the information of sample relations.
In the ablation studies, we show how the performance can be further improved with a simple batch construction strategy at test time. For more discussion on using MPN at test time, we refer the reader to the supplementary material.

%% file: tex_files/experiments.tex
\section{Experiments}

In this section, we compare our proposed approach to state-of-the-art deep metric learning approaches on four public benchmarks. To underline the effectiveness of our approach, we further present an extensive ablation study.
\vspace{-0.3cm}
\subsection{Implementation Details}

We implement our method in PyTorch \cite{Paszke17} library. 
Following other works \cite{DBLP:conf/iccv/BrattoliRO19,DBLP:conf/cvpr/Cakir0XKS19,DBLP:conf/iccv/ManmathaWSK17,DDBLP:conf/cvpr/Sanakoyeu2019,DBLP:journals/corr/abs-2004-01113,DBLP:conf/wacv/XuanSP20,DBLP:journals/corr/abs-1811-12649}, we present results using ResNet50 \cite{7780459} pretrained on ILSVRC 2012-CLS dataset \cite{DBLP:journals/corr/RussakovskyDSKSMHKKBBF14} as backbone CNN. Like the majority of recent methods \cite{DBLP:conf/eccv/GeHDS18, DBLP:conf/cvpr/KimKCK20,DBLP:conf/cvpr/Park2019, DBLP:journals/corr/abs-1909-05235,DDBLP:conf/cvpr/Wand2019, DBLP:conf/cvpr/WangHKHGR19, DBLP:journals/corr/abs-2010-13636}, we use embedding dimension of sizes $512$ for all our experiments and low temperature scaling for the softmax cross-entropy loss function \cite{DBLP:conf/icml/GuoPSW17}. Furthermore, we preprocess the images following \cite{DBLP:conf/cvpr/KimKCK20}. 
We resize the cropped image to $227 \times 227$, followed by applying a random horizontal flip.
During test time, we resize the images to $256 \times 256$ and take a center crop of size $227 \times 227$. 
We train all networks for $70$ epochs using RAdam optimizer \cite{DBLP:journals/corr/abs-1908-03265}.
To find all hyperparameters we perform random search \cite{DBLP:journals/jmlr/BergstraB12}. For mini-batch construction, we first randomly sample a given number of classes, followed by randomly sampling a given number of images for each class as commonly done in metric learning \cite{DBLP:conf/eccv/GrLoss,DBLP:conf/cvpr/SchroffKP15,DBLP:journals/corr/abs-2004-01113, DBLP:journals/corr/abs-1811-12649}. We use small mini-batches of size 50-100 and provide an analysis on different numbers of classes and samples on CUB-200-2011 and Cars196 in the supplementary. Our forward pass takes 73\% of time for the backbone and the remaining for the MPN. All the training is done in a single TitanX GPU, \textit{i.e.}, the method is memory efficient.
\vspace{-0.3cm}
\subsection{Benchmark Datasets and Evaluation Metrics}

\textbf{Datasets}: We conduct experiments on 4 publicly available datasets using the conventional splitting protocols \cite{DBLP:conf/cvpr/SongXJS16}:
\begin{itemize}
\item
CUB-200-2011 \cite{WahCUB_200_2011} consists of $200$ classes of birds with each class containing $58$ images on average. For training, we use the first $100$ classes and for testing the remaining classes.
\item
Cars196 \cite{KrauseStarkDengFei-Fei_3DRR2013} contains $196$ classes representing different cars with each class containing on average $82$ images. We use the first $98$ classes for training and the remaining classes for testing.
\item
Stanford Online Products (SOP) \cite{DBLP:conf/cvpr/SongXJS16} consists of 22,634 classes  ($5$ images per class on average) of product images from ebay. We use $11,318$ classes for training and the remaining $11,316$ classes for testing. 
\item
In-Shop Clothes \cite{DBLP:conf/cvpr/LiuLQWT16} contains $7,982$ classes of clothing items, with each class having $4$ images on average. We use $3,997$ classes for training, while the test set, containing $3,985$ classes, is split into a query set and a gallery set.
\end{itemize}

\textbf{Evaluation Metrics}: For evaluation, we use the two commonly used evaluation metrics, Recall@K (R@K) \cite{DBLP:journals/pami/JegouDS11} and Normalized Mutual Information (NMI) \cite{DBLP:journals/corr/abs-1110-2515}. The first one evaluates the retrieval performance by computing the percentage of images whose $K$ nearest neighbors contain at least one sample of the same class as the query image. To evaluate the clustering quality, we apply K-means clustering \cite{MacQueen} on the embedding feature vectors of all test samples, and compute NMI based on this clustering. To be more specific, NMI evaluates how much the knowledge about the ground truth classes increases given the clustering obtained by the K-means algorithm. 


\subsection{Comparison to state-of-the-art}
\label{subsec:ComparisonSOTA}

\input{tables/loss_all}
\input{tables/inshop}

\noindent\textbf{Quantitative Results.}
In Table~\ref{tab:res_loss}, we present the results of our method and compare
them with the results of other approaches on CUB-200-2011 \cite{WahCUB_200_2011}, Cars196 \cite{KrauseStarkDengFei-Fei_3DRR2013}, and Stanford Online Products \cite{DBLP:conf/cvpr/SongXJS16}. 
On CUB-200-2011 dataset, our method reaches $70.3$ Recall@1, an improvement of $0.6$ percentage points (pp) over the state-of-the-art \textit{Proxy Anchor} \cite{DBLP:conf/cvpr/KimKCK20} using ResNet50 backbone. On the NMI metric, we outperform the highest scoring method, \textit{DiVA} \cite{DBLP:conf/eccv/MilbichRBSBOC20} by $2.6pp$. On Cars196, we reach $88.1$ Recall@1, an improvement of $0.4pp$ over \textit{Proxy Anchor} \cite{DBLP:conf/cvpr/KimKCK20} with ResNet50 backbone. On the same dataset, we reach $74.8$ on the NMI score, $0.8pp$ higher than the previous best-performing method, \textit{Normalized Softmax} \cite{DBLP:journals/corr/abs-1811-12649}. On Stanford Online Products dataset, our method reaches $81.4$ Recall@1 which is $1.3pp$ better than the previous best method, \textit{HORDE} \cite{DBLP:journals/corr/abs-1908-02735}. On the NMI metric, our method reaches the highest score, outperforming \textit{SoftTriple Loss} \cite{DBLP:journals/corr/abs-1909-05235} by $0.6pp$. 

Finally, we present the results of our method on the In-Shop Clothes dataset in Table~\ref{tab:res_inshop}. 
Our method reaches $92.8$ Recall@1, an improvement of $0.7pp$ over the previous best method \textit{Proxy Anchor} \cite{DBLP:conf/cvpr/KimKCK20} with ResNet50 backbone. In summary, while in the past, different methods (\textit{Proxy Anchor} \cite{DBLP:conf/cvpr/KimKCK20}, \textit{ProxyNCA++} \cite{DBLP:journals/corr/abs-2004-01113}, \textit{Normalized Softmax} \cite{DBLP:journals/corr/abs-1811-12649}, \textit{HORDE} \cite{DBLP:journals/corr/abs-1908-02735}, \textit{SoftTriple Loss} \cite{DBLP:journals/corr/abs-1909-05235}, \textit{DiVA} \cite{DBLP:conf/eccv/MilbichRBSBOC20}) scored the highest in at-least one metric, now our method reaches the best results in all Recall@$1$ and NMI metrics across all four datasets. 

\begin{figure}[ht!]
\begin{center}
    \centerline{\includegraphics[width=.99\linewidth]{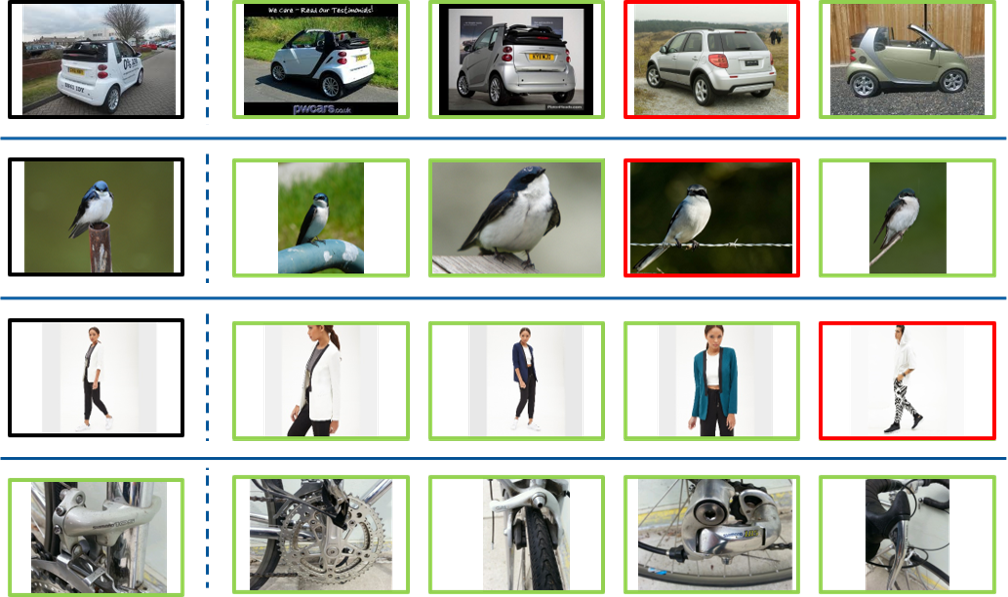}}
\caption{Retrieval results on a set of images from
\textit{CUB-200-2011} (top), \textit{Cars196} (second from top), \textit{Stanford Online Products} (second from bottom), and \textit{In-Shop Clothes} (bottom) datasets using our model. The most left column contains query images and the results are ranked by distance. Green frames indicate that the retrieved image is from the same class as the query image, while red frames indicate that the retrieved image is from a different class.}
\vspace{-1cm}
\label{fig:retrieval_new}
\end{center}
\end{figure}

\noindent\textbf{Qualitative Results.}
In Figure~\ref{fig:retrieval_new}, we present qualitative results on the retrieval task for all four datasets. In all cases, the query image is given on the left, with the four nearest neighbors given on the right. Green boxes indicate cases where the retrieved image is of the same class as the query image, and red boxes indicate a different class. In supplementary material, we provide qualitative evaluations on the clustering performance using t-SNE \cite{DBLP:journals/ml/MaatenH12} visualization.

\begin{figure*}[thp!]
\centering
\begin{minipage}{.34\textwidth}
  \begin{center}
  \centerline{\includegraphics[width=1\linewidth]{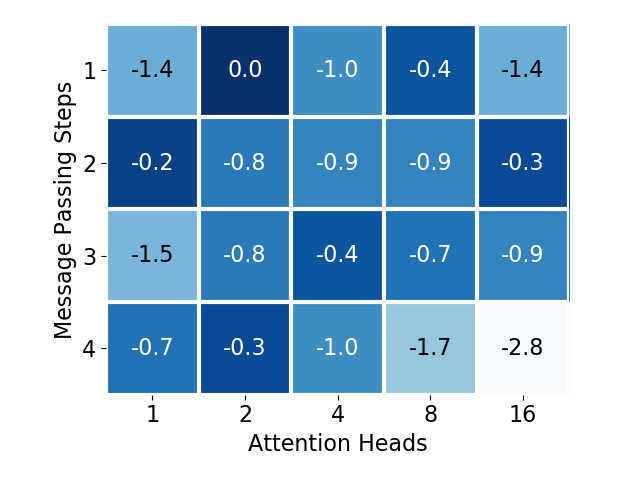}}
  \caption{Relative difference to the best model with respect to Recall@1 on CUB-200-2011.} \label{fig:cub_LH}
  \end{center}
\end{minipage}%
\hfill
\hspace{0.01cm}
\begin{minipage}{.34\textwidth}
  \begin{center}
  \centerline{\includegraphics[width=1\linewidth]{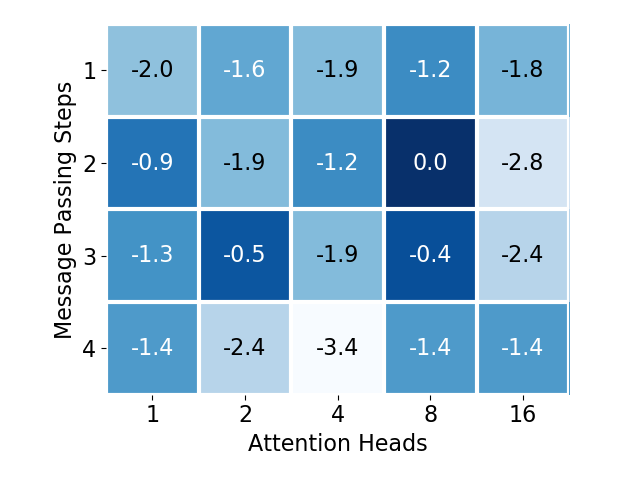}}
  \caption{Relative difference to the best \newline model with respect to Recall@1 on Cars196.} \label{fig:cars_LH}
  \end{center}
\end{minipage}%
\hfill
\begin{minipage}{.31\textwidth}
  \begin{center}
  \centerline{\includegraphics[width=1\linewidth]{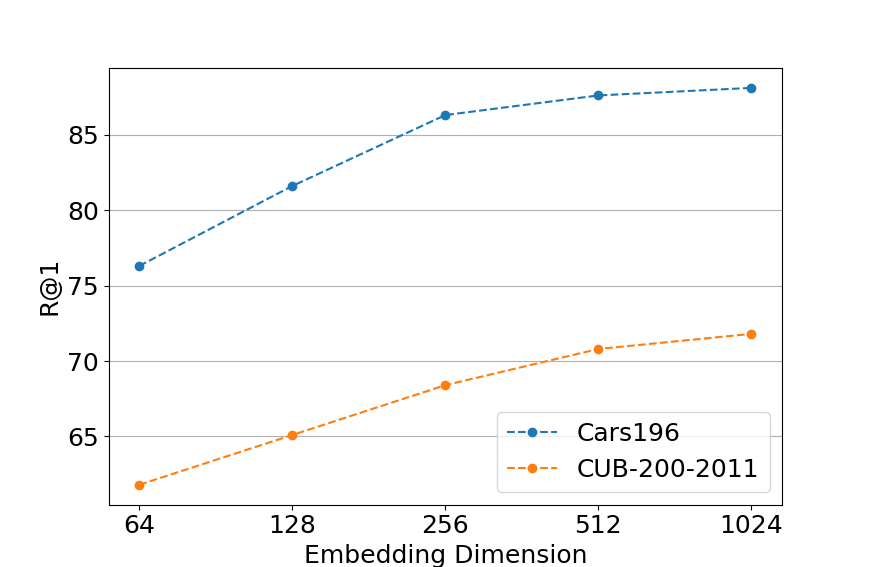}}
  \vspace{0.5cm}
  \caption{Performance for different embedding dimensions on CUB-200-2011 and Cars196.} \label{fig:cars_embed}
  \end{center}
\end{minipage}
\vspace{0.3cm}
\end{figure*}

\vspace{-0.2cm}
\subsection{Ablation Studies and Robustness Analysis}
\label{subsec:ablations}

In this section, we use the CUB-200-2011 \cite{WahCUB_200_2011} and Cars196 \cite{KrauseStarkDengFei-Fei_3DRR2013} datasets to analyze the robustness of our method and show the importance of our design choices.

\noindent\textbf{MPN Matters.} To show the performance improvement when using the MPN during training, we conduct experiments by training the backbone CNN solely with the auxiliary loss, i.e., the cross-entropy loss on the backbone CNN, and without MPN (see the first row in Table~\ref{tab:w_MPN_during_test_time}). For a fair comparison, we use the same implementation details as for the training with MPN. On CUB-200-2011, this leads to a performance drop of $2.8pp$ in Recall@1 (to $67.5$) and $4.2pp$ in NMI (to $69.8$). On Cars196, it leads to a more significant performance drop of $3.9pp$ in Recall@1 (to $84.2$) and $6.1pp$ in NMI (to $68.7$), showing the benefit of our proposed formulation. 

\begin{figure}
    \centering
    \includegraphics[width=0.5\textwidth]{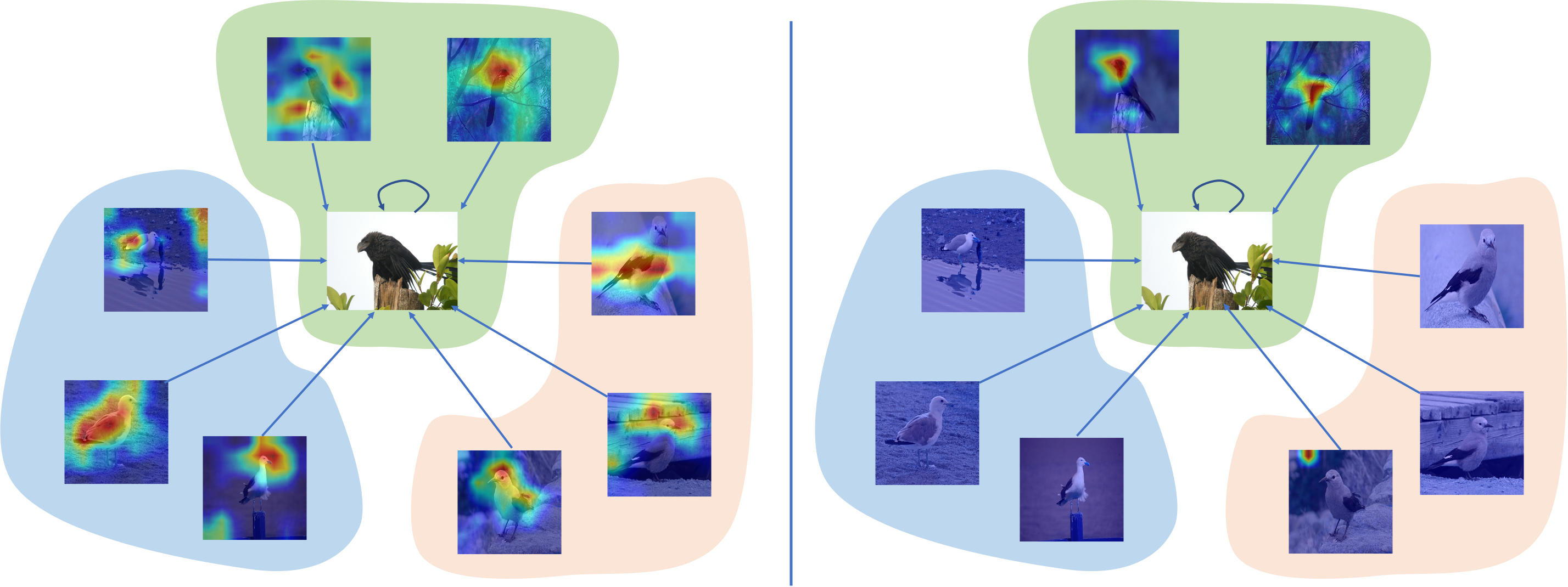}
    \caption{Comparison of the embeddings of a given batch after one epoch of training without and with MPN.}
    \label{fig:trained_untrained}
\end{figure}

To give an intuition of how the MPN evolves during the training process, we use GradCam \cite{DBLP:journals/ijcv/SelvarajuCDVPB20} to observe which neighbors a sample relies on when computing the final class prediction after the MPN \cite{DBLP:journals/ijcv/SelvarajuCDVPB20}. To do so, we compare the predictions of an untrained MPN to a trained one. As can be seen in the left part of Figure~\ref{fig:trained_untrained}, the untrained MPN takes information from nearly all samples in the batch into account, where red, blue, and green represent different classes. The trained MPN (left part of Figure~\ref{fig:trained_untrained}) only relies on the information of samples of the same class. This suggests that using the MPN with self-attention scores as edge weights enforces the embeddings of negative and positive samples to become more dissimilar and similar, respectively. In supplementary, we also provide and compare visualizations of the embedding vectors of a batch of samples after one epoch of training and of all test samples after the whole training.

\noindent\textbf{Number of Message Passing Steps and Attention Heads.} In Figure~\ref{fig:cub_LH}, we investigate the robustness of the algorithm when we differ the number of message passing steps and attention heads of our MPN. On CUB-200-2011 dataset, we reach the best results when we use a single message passing step, containing two attention heads. We see that increasing the number of message passing steps or the number of attention heads, for the most part, does not result in a large drop in performance. The biggest drop in performance happens when we use four message-passing steps, each having sixteen attention heads. 
In Figure~\ref{fig:cars_LH}, we do a similar robustness analysis for the Cars196 dataset. Unlike CUB-200-2011, the method performs best using two layers and eight attention heads. However, it again performs worst using four message passing steps.
This observation is in line with \cite{DBLP:conf/iclr/VelickovicCCRLB18}, which also utilizes a few message passing steps when applying graph attention. 

\noindent\textbf{Embedding Dimension.} In Figure~\ref{fig:cars_embed}, we measure the performance of the model as a function of the embedding size. We observe that the performance of the network increases on both datasets when we increase the size of the embedding layer. This is unlike \cite{DDBLP:conf/cvpr/Wand2019}, which reports a drop in performance when the size of the embedding layer gets bigger than $512$. While increasing the dimension of the embedding layer results in even better performance, for fairness with the other methods that do not use an embedding size larger than $512$, we avoid those comparisons. 

\noindent\textbf{Auxiliary Loss Function.} Considering that in the default scenario, we do not use the MPN during inference, we investigate the effect of adding the auxiliary loss function at the top of the backbone CNN embedding layer. 
On CUB-200-2011 dataset, we see that such a loss helps the network improve by $2.2pp$ in Recall@1. Without the loss, the performance of the network drops to $68.1$ as shown in the second row of Table~\ref{tab:w_MPN_during_test_time}. On the other hand, removing the auxiliary loss function leads to a performance drop of only $0.9pp$ in Recall@1 on Cars196 to $87.2$. However, the NMI performance drops by $2.7pp$ to $72.1$ on Cars196 and $2.0pp$ on CUB-200-2011. 
%

\begin{figure}[htb]
    \vspace{-0.3cm}
    \begin{center}
    \centerline{\includegraphics[width=1\linewidth]{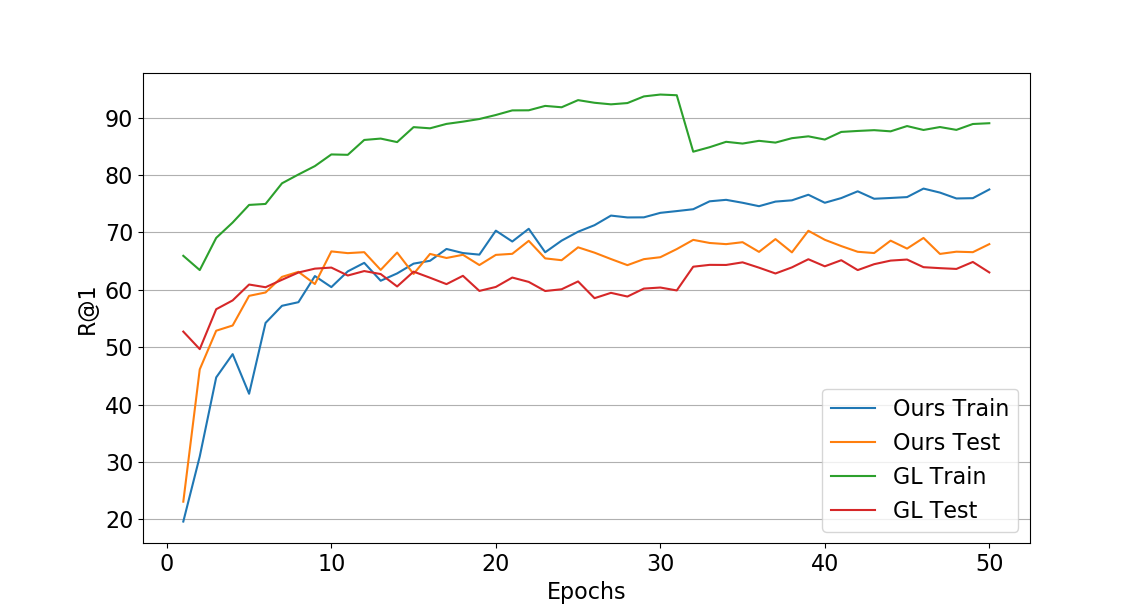}}
    \caption{Performance on training and test data of CUB-200-2011 compared to Group Loss \cite{DBLP:conf/eccv/GrLoss}.}
    \label{fig:cub_train_test}
    \end{center}
    \vspace{-0.2cm}
\end{figure}

\noindent\textbf{Implicit Regularization.} We further investigate the training behavior of our proposed approach on CUB-200-2011. As already stated above, Group Loss \cite{DBLP:conf/eccv/GrLoss} also utilized contextual classification, with the authors claiming that it introduces implicit regularization and thus less overfitting. However, their approach is based on a hand-crafted label propagation rule, while ours takes into account the contextual information in an end-to-end learnable way. Therefore, we present the training behavior of our approach and compare it to the behavior of the Group Loss \cite{DBLP:conf/eccv/GrLoss}.
As can be seen in Figure~\ref{fig:cub_train_test}, Group Loss \cite{DBLP:conf/eccv/GrLoss} shows higher overfitting on the training data, while our method is capable of better generalization on the test dataset and has a smaller gap between training and test performance. We argue that by taking into account the global structure of the dataset in an end-to-end learnable way, our approach is able to induce an even stronger implicit regularization.

\noindent\textbf{Using MPN During Test Time.} In Table~\ref{tab:w_MPN_during_test_time}, we analyze the effect of applying message passing during inference (see row four). On CUB-200-2011 dataset, we improve by $0.5pp$ in Recall@1, and by $0.5pp$ on the NMI metric. On Cars196 dataset, we also gain $0.5pp$ in Recall@1 by using MPN during inference. More impressively, we gain $1.4pp$ in the NMI metric, putting our results $2.2pp$ higher than \textit{Normalized Softmax} \cite{DBLP:journals/corr/abs-1811-12649}. We gain an improvement in performance in all cases, at the cost of extra parameters.

Note, our method does not require the usage of these extra parameters in inference. As we have shown, for a fair comparison, our method reaches state-of-the-art results even without using MPN during inference (see Tables~\ref{tab:res_loss} and \ref{tab:res_inshop}). We consider the usage of MPN during inference a performance boost, but not a central part of our work.

\input{tables/MPN}
\noindent\textbf{Ensembles.} The Group Loss \cite{DBLP:conf/eccv/GrLoss} showed that the performance of their method significantly improves by using an ensemble at test time. The ensemble was built by simply concatenating the features of $k$ independently trained networks. Similarly, we also conduct experiments on ensembles using $2$ and $5$ networks, respectively, and compare our ensemble with that of \cite{DBLP:conf/eccv/GrLoss}.
\input{tables/ensembles}

In Table~\ref{tab:ensembles}, we present the results of our ensembles. We see that when we use $2$ networks, the performance increases by $1.9pp$ on CUB-200-2011, $3.0pp$ on Cars196, and $0.4pp$ on Stanford Online Products.  Similarly, the NMI score also improves by $0.3pp$ on CUB-200-2011, $0.1pp$ on Cars196, and $0.1pp$ on Stanford Online Products. Unfortunately, the Recall@$1$ performance on In-Shop Clothes only improves by $0.1pp$.
Using $5$ networks, the performance increases by $2.8pp$ on CUB-200-2011, $3.4pp$ on Cars196, $0.7pp$ on Stanford Online Products, and $0.6pp$ on In-Shop Clothes compared to using a single network. NMI on CUB-200-2011 is improved by $0.4pp$ compared to a single network, on Cars196 it increases by $0.6pp$ more and on Stanford Online Products it increases by $0.6pp$.

Compared to \cite{DBLP:conf/eccv/GrLoss}, the performance increase of our approach from one network to an ensemble is higher. This is surprising, considering that our network starts from a higher point, and has less room for improvement.
\noindent

%% file: tables/loss_all.tex
\begin{table*}
\centering
    \resizebox{\textwidth}{!}{%
    \begin{tabular}{@{}l|c|ccccc|ccccc|cccc@{}}

\hline
\textbf{} & & \multicolumn{5}{c}{CUB-200-2011} & \multicolumn{5}{c}{CARS196} & \multicolumn{4}{c}{Stanford Online Products}\\ \hline
\textbf{Method} & BB & \textbf{R@1} & \textbf{R@2} & \textbf{R@4} & \textbf{R@8} & \textbf{NMI} &  \textbf{R@1} & \textbf{R@2} & \textbf{R@4} & \textbf{R@8} & \textbf{NMI} & \textbf{R@1} & \textbf{R@10} & \textbf{R@100} & \textbf{NMI} \\ \hline
Triplet$^{64}$ \cite{DBLP:conf/cvpr/SchroffKP15} \textit{CVPR15} & G & 42.5 & 55 & 66.4 & 77.2 & 55.3 & 51.5 & 63.8 & 73.5 & 82.4 & 53.4 & 66.7 & 82.4 & 91.9 & 89.5 \\

Npairs$^{64}$ \cite{DBLP:conf/nips/Sohn16} \textit{NeurIPS16} & G & 51.9 & 64.3 & 74.9 & 83.2 & 60.2 &  68.9 & 78.9 & 85.8 & 90.9 & 62.7 &  66.4 & 82.9 & 92.1 & 87.9  \\

Deep Spectral$^{512}$ \cite{DBLP:conf/icml/LawUZ17} \textit{ICML17} & BNI & 53.2 & 66.1 & 76.7 & 85.2 & 59.2 &  73.1 & 82.2 & 89.0 & 93.0 & 64.3 &  67.6 & 83.7 & 93.3& 89.4 \\ 

Angular Loss$^{512}$ \cite{DBLP:conf/iccv/WangZWLL17} \textit{ICCV17} & G & 54.7 & 66.3 & 76 & 83.9 & 61.1 &  71.4 & 81.4 & 87.5 & 92.1 & 63.2 &  70.9 & 85.0 & 93.5 & 88.6\\

Proxy-NCA$^{64}$ \cite{DBLP:conf/iccv/Movshovitz-Attias17} \textit{ICCV17} & BNI & 49.2 & 61.9 & 67.9 & 72.4 & 59.5 &  73.2 & 82.4 & 86.4 & 88.7 & 64.9 &  73.7 & - & - & 90.6 \\

Margin Loss$^{128}$  \cite{DBLP:conf/iccv/ManmathaWSK17} \textit{ICCV17} & R50 & 63.6 & 74.4 & 83.1 & 90.0 & 69.0 &  79.6 & 86.5 & 91.9 & 95.1 & 69.1 &  72.7 & 86.2 & 93.8 & 90.7 \\

Hierarchical triplet$^{512}$  \cite{DBLP:conf/eccv/GeHDS18} \textit{ECCV18} & BNI & 57.1 & 68.8 & 78.7 & 86.5 & - &  81.4 & 88.0 & 92.7 & 95.7 &  - & 74.8 & 88.3 & 94.8 & - \\

ABE$^{512}$  \cite{DBLP:conf/eccv/KimGCLK18} \textit{ECCV18} & G & 60.6 & 71.5 & 79.8 & 87.4 & - & 85.2 & 90.5 & 94.0 & 96.1 &  - & 76.3 & 88.4 &  94.8 & - \\

Normalized Softmax$^{512}$ \cite{DBLP:journals/corr/abs-1811-12649} \textit{BMVC19} & R50 & 61.3 & 73.9 & 83.5 & 90.0 & 69.7 &  84.2 & 90.4 & 94.4 & 96.9 & \textbf{\textcolor{red}{74.0}} & 78.2 & 90.6 & 96.2 & 91.0 \\

RLL-H$^{512}$  \cite{DBLP:conf/cvpr/WangHKHGR19} \textit{CVPR19} & BNI & 57.4 & 69.7 & 79.2 & 86.9 & 63.6 & 74 & 83.6 & 90.1 & 94.1 &   65.4 &   76.1 & 89.1 & 95.4 & 89.7 \\

Multi-similarity$^{512}$  \cite{DDBLP:conf/cvpr/Wand2019} \textit{CVPR19} & BNI & 65.7 & 77.0 & 86.3 & 91.2 & - &  84.1 & 90.4 & 94.0 & 96.5 & - & 78.2 & 90.5 & 96.0 & - \\ 

Relational Knowledge$^{512}$ \cite{DBLP:conf/cvpr/Park2019} \textit{CVPR19} & G & 61.4 & 73.0 & 81.9 & 89.0 & - &  82.3 & 89.8 & 94.2 & 96.6 & - & 75.1 & 88.3 & 95.2  & -\\ 

Divide and Conquer$^{1028}$  \cite{DDBLP:conf/cvpr/Sanakoyeu2019} \textit{CVPR19} & R50 & 65.9 & 76.6 & 84.4 & 90.6 & 69.6 & 84.6 & 90.7 & 94.1 & 96.5 &  70.3 &   75.9   & 88.4 & 94.9 &90.2 \\ 


SoftTriple Loss$^{512}$ \cite{DBLP:journals/corr/abs-1909-05235} \textit{ICCV19} & BNI & 65.4 & 76.4 & 84.5 & 90.4 & 69.3 & 84.5 & 90.7 & 94.5 & 96.9 & 70.1 & 78.3 & 90.3 & 95.9 & \textbf{\textcolor{red}{92.0}} \\

HORDE$^{512}$ \cite{DBLP:journals/corr/abs-1908-02735} \textit{ICCV19} & BNI & 66.3 & 76.7 & 84.7 & 90.6 & -  & 83.9 & 90.3 & 94.1 & 96.3 & -  & \textbf{\textcolor{red}{80.1}} & \textbf{\textcolor{red}{91.3}} & 96.2 & -\\

MIC$^{128}$ \cite{DBLP:conf/iccv/BrattoliRO19} \textit{ICCV19} & R50 & 66.1 & 76.8 & 85.6 & - & 69.7 & 82.6 & 89.1 & 93.2 & - & 68.4 & 77.2 & 89.4 & 95.6 & 90.0  \\

Easy triplet mining$^{512}$ \cite{DBLP:conf/wacv/XuanSP20} \textit{WACV20} & R50 & 64.9 & 75.3 & 83.5 & - & - & 82.7 & 89.3 & 93.0 & - & - & 78.3 & 90.7 & \textbf{\textcolor{red}{96.3}} & - \\   

Group Loss$^{1024}$ \cite{DBLP:conf/eccv/GrLoss} \textit{ECCV20} & BNI & 65.5 & 77.0 & 85.0 & 91.3 & 69.0 & 85.6 & 91.2 & 94.9 & 97.0 & \textbf{\textcolor{blue}{72.7}} & 75.1 & 87.5 & 94.2 & \textbf{\textcolor{blue}{90.8}} \\

Proxy NCA++$^{512}$ \cite{DBLP:journals/corr/abs-2004-01113} \textit{ECCV20} & R50 & 66.3 & 77.8 & \textbf{\textcolor{red}{87.7}} & 91.3 & \textbf{\textcolor{blue}{71.3}} & 84.9 & 90.6 & 94.9 & 97.2 & 71.5 & 79.8 & \textbf{\textcolor{red}{91.4}} & \textbf{96.4} & - \\

DiVA$^{512}$ \cite{DBLP:conf/eccv/MilbichRBSBOC20} \textit{ECCV20} & R50 & \textbf{\textcolor{blue}{69.2}} & \textbf{\textcolor{blue}{79.3}} & - & - & \textbf{\textcolor{red}{71.4}} & \textbf{\textcolor{blue}{87.6}}  & \textbf{\textcolor{red}{92.9}} & - & - & 72.2 & 79.6 & - & - & 90.6 \\

PADS$^{128}$ \cite{DBLP:conf/cvpr/RothMO20} \textit{CVPR20} & R50 & 67.3  & 78.0  & 85.9 & - & 69.9 & 83.5 & 89.7 & 93.8 & - & 68.8 & 76.5  & 89.0 &  95.4 & 89.9 \\

Proxy Anchor$^{512}$ \cite{DBLP:conf/cvpr/KimKCK20} \textit{CVPR20} & BNI & 68.4 & 79.2 & 86.8 & \textbf{\textcolor{blue}{91.6}} & -  & 86.1 & 91.7 & 95.0 & \textbf{\textcolor{blue}{97.3}} & -  & 79.1 & 90.8 & 96.2 & - \\

Proxy Anchor$^{512}$ \cite{DBLP:conf/cvpr/KimKCK20} \textit{CVPR20} & R50 & \textbf{\textcolor{red}{69.7}} & \textbf{\textcolor{red}{80.0}} & \textbf{\textcolor{blue}{87.0}} & \textbf{\textcolor{red}{92.4}} & -  & \textbf{\textcolor{red}{87.7}} & \textbf{\textcolor{red}{92.9}} & \textbf{\textcolor{red}{95.8}} & \textbf{\textcolor{red}{97.9}} & -  & \textbf{\textcolor{blue}{80.0}} & \textbf{91.7} & \textbf{96.6} & - \\

Proxy Few$^{512}$ \cite{DBLP:journals/corr/abs-2010-13636} \textit{NeurIPS20} & BNI & 66.6 & 77.6 & 86.4 & - & 69.8 & 85.5 & \textbf{\textcolor{blue}{91.8}} & \textbf{\textcolor{blue}{95.3}} & - & 72.4 & 78.0 & 90.6 & 96.2 & 90.2 \\ \hline





Ours$^{512}$ & R50 & \textbf{70.3} & \textbf{80.3} & \textbf{87.6} & \textbf{92.7} & \textbf{74.0} & \textbf{88.1} & \textbf{93.3} & \textbf{96.2} & \textbf{98.2} & \textbf{74.8} & \textbf{81.4} & \textbf{\textcolor{red}{91.3}} & 95.9 & \textbf{92.6} \\ \hline


\end{tabular}}

\caption{Retrieval and Clustering performance on \textit{CUB-200-2011}, \textit{CARS196} and \textit{Stanford Online Products} datasets. Bold indicates best, red second best, and blue third best results. The exponents attached to the method name indicates the embedding dimension. BB=backbone, G=GoogLeNet, BNI=BN-Inception and R50=ResNet50.}
\vspace{+0.5cm}
\label{tab:res_loss}

\end{table*}

%% file: tables/inshop.tex
\begin{table}[!ht]
\centering
\resizebox{0.49\textwidth}{!}{%
\begin{tabular}{@{}l|c|cccc@{}}
\toprule
\textbf{Method} & BB &\textbf{R@1} & \textbf{R@10} & \textbf{R@20} & \textbf{R@40} \\ \hline
FashionNet$^{4096}$ \cite{DBLP:conf/cvpr/LiuLQWT16} \textit{CVPR16} & V & 53.0 & 73.0 & 76.0 & 79.0 \\
A-BIER$^{512}$ \cite{DBLP:journals/pami/OpitzWPB20} \textit{PAMI20} & G & 83.1 & 95.1 & 96.9 & 97.8 \\
ABE$^{512}$ \cite{DBLP:conf/eccv/KimGCLK18} \textit{ECCV18} & G & 87.3 & 96.7 & 97.9 & 98.5 \\
Multi-similarity$^{512}$  \cite{DDBLP:conf/cvpr/Wand2019} \textit{CVPR19} & BNI & 89.7 & \textbf{\textcolor{blue}{97.9}} & 98.5 & \textbf{\textcolor{red}{99.1}} \\
Learning to Rank$^{512}$ \cite{DBLP:conf/cvpr/Cakir0XKS19} & R50 & 90.9 & 97.7 & 98.5 & \textbf{\textcolor{blue}{98.9}} \\
HORDE$^{512}$ \cite{DBLP:journals/corr/abs-1908-02735} \textit{ICCV19} & BNI & 90.4 & 97.8 & 98.4 & \textbf{\textcolor{blue}{98.9}} \\
MIC$^{128}$ \cite{DBLP:conf/iccv/BrattoliRO19} \textit{ICCV19} & R50 & 88.2 & 97.0 & 98.0 & 98.8 \\
Proxy NCA++$^{512}$ \cite{DBLP:journals/corr/abs-2004-01113} \textit{ECCV20} & R50 & 90.4 & \textbf{\textcolor{red}{98.1}} & \textbf{\textcolor{red}{98.8}}  & \textbf{99.2} \\
Proxy Anchor$^{512}$ \cite{DBLP:conf/cvpr/KimKCK20} \textit{CVPR20} & BNI & \textbf{\textcolor{blue}{91.5}} & \textbf{\textcolor{red}{98.1}} & \textbf{\textcolor{red}{98.8}} & \textbf{\textcolor{red}{99.1}} \\ 
Proxy Anchor$^{512}$ \cite{DBLP:conf/cvpr/KimKCK20} \textit{CVPR20} & R50 & \textbf{\textcolor{red}{92.1}} & \textbf{\textcolor{red}{98.1}} & \textbf{\textcolor{blue}{98.7}} & \textbf{99.2}\\ \hline
Ours$^{512}$ & R50 & \textbf{92.8} & \textbf{98.5} & \textbf{99.1} & \textbf{99.2} \\ \hline

\end{tabular}%
}
\caption{Retrieval performance on \textit{In Shop Clothes}. }
\vspace{-0.5cm}
\label{tab:res_inshop}
\end{table}

%% file: tables/MPN.tex
\begin{table}[hbt!]
\tiny
\centering
\resizebox{0.49\textwidth}{!}{
\begin{tabular}{@{}l|c|cc|cc|cc@{}}

& & \multicolumn{2}{c}{CUB-200-2011} & \multicolumn{2}{c}{CARS196} \\ 
\hline
Training Losses & Test Time Embeddings & \textbf{R@1} &  \textbf{NMI} & \textbf{R@1} &  \textbf{NMI}\\
\hline
Cross-Entropy & Backbone Embeddings & 67.5 & 69.8 & 84.2 & 68.7 \\
MPN Loss & Backbone Embeddings & 68.1 & 72.0 & 87.2 & 72.1\\
MPN Loss + Auxiliary Loss & Backbone Embeddings  & 70.3 & 74.0 & 88.1 & 74.8 \\
MPN Loss + Auxiliary Loss & MPN Embeddings & \textbf{70.8} & \textbf{74.5} & \textbf{88.6} & \textbf{76.2} \\
\hline
\end{tabular}}
\caption{Performance of the network with and without MPN during training and testing time. We achieved all results using embedding dimension 512.}
\label{tab:w_MPN_during_test_time}
\end{table}

%% file: tables/ensembles.tex
\begin{table}[hbt!]
\centering
\resizebox{0.49\textwidth}{!}{
\begin{tabular}{@{}l|cc|cc|cc|cc@{}}
\hline
& \multicolumn{2}{c}{CUB-200-2011} & \multicolumn{2}{c}{Cars196} & \multicolumn{2}{c}{Stanford Online Products} & \multicolumn{1}{c}{In-Shop Clothes}\\ 
\hline
 & \textbf{R@1} &  \textbf{NMI} & \textbf{R@1} &  \textbf{NMI} &
 \textbf{R@1} &  \textbf{NMI}& \textbf{R@1} \\
\hline
GL & 65.5 & 69.0 & 85.6 & 72.7 & 75.7 & 91.1 & - \\
Ours & 70.3 & 74.0 & 88.1 & 74.8 & 81.4 & 92.6 & 92.8 \\
\hline
GL 2 & 65.8 & 68.5 & 86.2 & 72.6 & 75.9 & 91.1 & - \\
Ours 2 & 72.2 & 74.3 & 90.9 & 74.9 & 81.8 & 92.7 & 92.9 \\
\hline
GL 5 & 66.9 & 70.0 & 88.0 & 74.2 & 76.3 & 91.1 & - \\
Ours 5 & \textbf{73.1} & \textbf{74.4} & \textbf{91.5} & \textbf{75.4} & \textbf{82.1} & \textbf{92.8} & \textbf{93.4} \\
\hline
\end{tabular}}
\caption{Performance of our ensembles and comparisons with the ensemble models of \cite{DBLP:conf/eccv/GrLoss}.}
\label{tab:ensembles}
\end{table}

%% file: tex_files/conclusion.tex
\vspace{0.2cm}
\section{Conclusions}
\vspace{0.2cm}

In this work, we propose a model that utilizes the power of message passing networks for the task of deep metric learning. Unlike classical metric learning methods, e.g., triplet loss, our model utilizes all the intra-batch relations in the mini-batch to promote similar embeddings for images coming from the same class, and dissimilar embeddings for samples coming from different classes. Our model is fully learnable, end-to-end trainable, and does not utilize any handcrafted rules. Furthermore, our model achieves state-of-the-art results while using the same number of parameters, and compute time, during inference. In future work, we will explore the applicability of our model for the tasks of semi-supervised deep metric learning and deep metric learning in the presence of only relative labels. 

\textbf{Acknowledgements.} This research was partially funded by the Humboldt Foundation through the Sofia Kovalevskaja Award. We thank Guillem Bras\'{o} for useful discussions.

%% file: tex_files/appendix.tex
\appendix


\section{Different Embedding Sizes}
\label{sec:embeddings}
\input{tables/embedding_size}

To investigate the robustness of our approach concerning different embedding sizes, we present the performance for several embedding dimensions on Stanford Online Products and In Shop Clothes, which complement the results presented on CUB-200-2011 and Cars196 in the main paper (see Figure 6 in the main paper). The results with embedding dimension up to $1024$ can be found in Table~\ref{tab:embedding_dim}. 
As one can see, the performance on the smaller datasets increases with increasing embedding dimension similar to Proxy Anchor \cite{DBLP:conf/cvpr/KimKCK20} while the performance on Stanford Online Products remains the same. On the other hand, similar to the Multi-Similarity loss \cite{DDBLP:conf/cvpr/Wand2019} the performance is shown to decrease on the In-Shop dataset if the size of the embedding layer becomes larger than $512$.

\section{Large Images and ProxyNCA++}
\label{sec:largeimages}
\input{tables/proxynca++}
\input{tables/larger_images}
Different from most approaches in the field of deep metric learning, \cite{DBLP:journals/corr/abs-2004-01113,DBLP:journals/corr/abs-1908-02735} crop the images during training to size $256 \times 256$, while at test time they first resize to $288 \times 288$ before again cropping to $256 \times 256$. Naturally, larger images are expected to lead to an increased performance. In the main work, for fair comparison, we report the performance of \cite{DBLP:journals/corr/abs-2004-01113} on the typical image size, i.e., $227 \times 227$. To obtain these numbers we ran their provided code\footnote{\url{https://github.com/euwern/proxynca_pp}} and compared the results to the Recall@$1$ given in their supplementary to validate the correctness, see Table~\ref{tab:proxynca++}.

We also evaluate the performance of our approach on images of size $256 \times 256$ to show that our performance also increases when using larger images. As can be seen in Table~\ref{tab:larger_img}, when we use larger images, our performance increases by $1.4pp$ Recall@$1$ and $0.3pp$ NMI on CUB-200-2011 and $1.9pp$ Recall@$1$ and $0.6pp$ NMI on Cars196. This leads to a even larger increase in performance compared to \cite{DBLP:journals/corr/abs-2004-01113,DBLP:journals/corr/abs-1908-02735}. 
We also outperform \cite{DBLP:conf/cvpr/KimKCK20} who also gave additional results for larger-size images (we already showed in the main paper that we outperform \cite{DBLP:conf/cvpr/KimKCK20} on regular-sized images). 

\section{Different Settings during Test Time}
\label{sec:MPNduringTestTime}
%
In this section, we detail three methods for batch construction when using the MPN during inference, as well as a teacher-student approach to avoid batch construction at test time.

%
%

\subsection{Batch Construction Based on Clustering}
\label{subsec:test_batches}
Ideally, we imitate the batch construction process that happens during training, i.e., sampling $n$ classes and taking $k$ samples for each. As we can not access ground truth labels during test time, we are not able to construct batches in this manner.

To this end, we first use randomly sampled batches to generate initial feature vectors using the backbone CNN. We then use an approximation of the ground truth class assignment by using a clustering algorithm. We compare $6$ common clustering algorithms as can be seen in Table~\ref{tab:settings}. 
Based on these clusters, we construct batches by sampling from $n$ clusters, $k$ samples each, analogous to the training procedure. This way, we ensure that every sample in the mini-batch communicates with samples from its own cluster (similar) and other clusters (dissimilar). We call the $k$ samples belonging to one cluster a {\it chunk}.
Finally, we compute refined feature vectors using MPNs and use those features for retrieval and clustering.

In general, clustering algorithms can be divided into two groups: the ones that require a fixed number of clusters to be generated, and the density-based cluster algorithms that need a minimum number of samples per cluster as well as a maximum distance $\epsilon$ between two samples to be considered as neighbors.

As in theory, we do not know the ground truth number of clusters during test time, we cannot use it for the cluster construction in the first group. Therefore, we conduct experiments on several different numbers of clusters and report the results on the number that performs best. To be specific, the algorithms achieve the best performance if we set the number of clusters to $900$. This number is significantly larger than the number of ground truth classes which is $100$ and $98$ for CUB-200-2011 and Cars196, respectively. 
Intuitively, using at least as many clusters as the number of classes is necessary since otherwise samples of different classes will be assigned to the same clusters. As \textit{overclustering} constructs more clusters than the ground truth number of classes and, therefore, smaller clusters than the ground truth class sizes, the cluster assignment is less prone to outliers. 
The performance of the algorithms that need a fixed number of classes can be seen in Table~\ref{tab:settings}, indicated by $\dagger$. However, they do not lead to a performance increase compared to the performance using solely the backbone architecture.


While we can bypass the oversampling issue by using the density-based algorithms (indicated by the $^*$ in Table~\ref{tab:settings}), none of the used clustering algorithms assigns clusters in a sufficiently accurate way such that the performance of the backbone CNN gets improved. 

\input{tables/eval_settings}

\subsection{Batch Construction Based on Nearest Neighbors}
As another option, we sample chunks by randomly choosing one \textit{anchor} and finding its $k-1$ nearest neighbors. Then we construct batches consisting of $n$ of these chunks and feed them through the MPN. In every batch, we only update the feature vectors of the anchors, meaning that we build one chunk for each image in the test set. As these chunks again can be highly noisy, the performance after the MPN drops compared to simply taking the embeddings from the backbone (see "nearest neighbors" in Table~\ref{tab:settings}).

\subsection{Reciprocal-kNN Batch Construction}
\begin{figure}[t]
\begin{center}
\centerline{\includegraphics[width=0.8\linewidth]{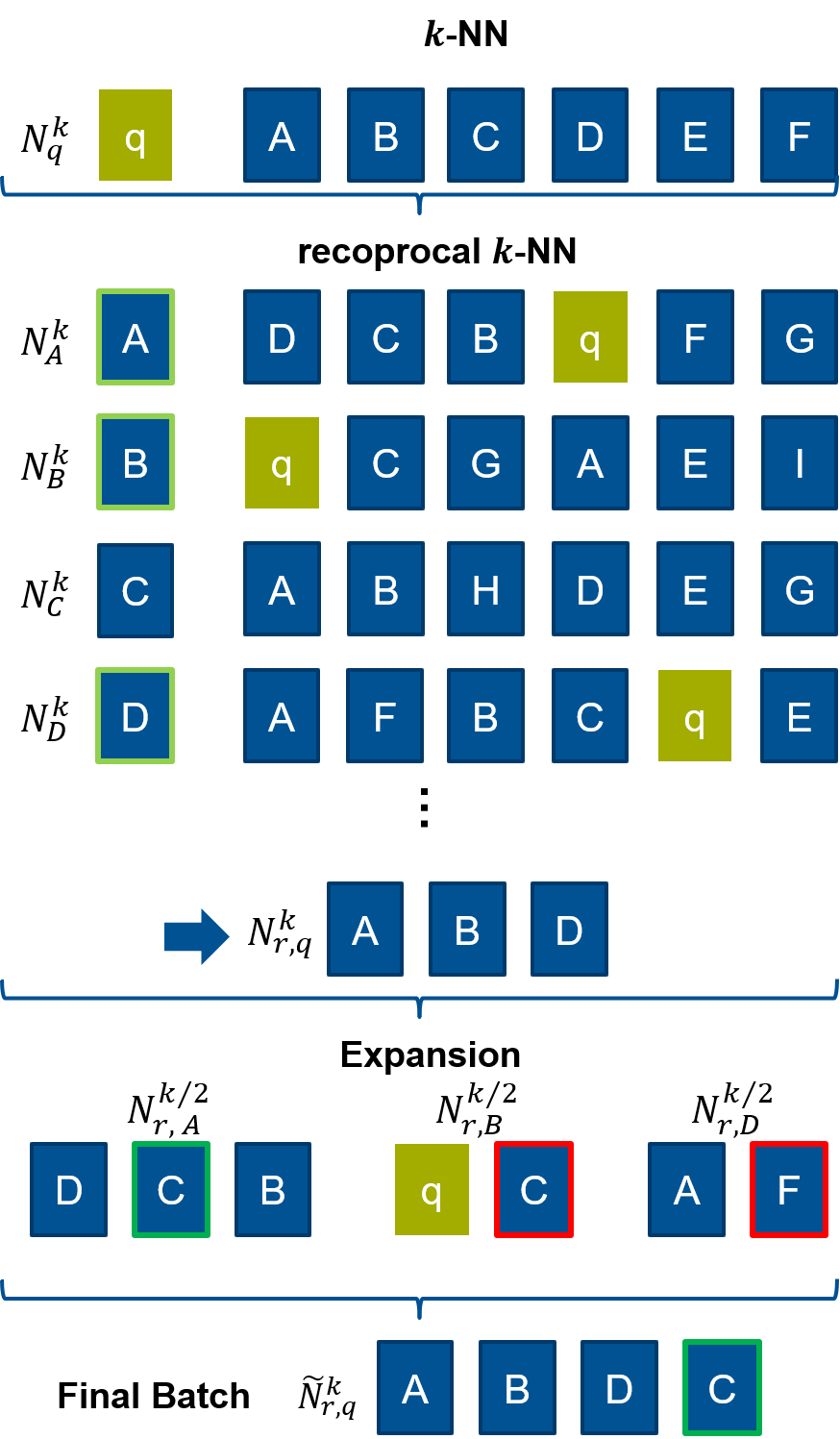}}
   \caption{Reciprocal k-nearest neighbor batch sampling for MPN during inference.}
   \end{center}
\label{fig:kreciprocal}
\end{figure}

Since imitating the training batch construction during test time and simply using a sample's $k-1$ nearest neighbors does not lead to a performance increase, we propose to construct batches more strictly during test time. 
To that end, we suggest constructing reciprocal $k$-nearest neighbor batches inspired by \cite{DBLP:conf/cvpr/ZhongZCL17}. Different from \cite{DBLP:conf/cvpr/ZhongZCL17} who use reciprocal $k$-nearest neighbor for evaluation, we use a similar idea for batch construction (we still do the final evaluation regularly, by simply evaluating Recall@K and NMI).
Knowing that samples that are highly similar to the query sample are more likely to be of the same class as the query image $c_q$ than dissimilar samples, we first compute the $k$-nearest neighbor set $N_q^k$ of a given query $q$ (see the upper part in Figure~\ref{fig:kreciprocal}). 
However, $N_q^k$ still might contain noisy samples. Therefore, we reduce $N_q^k$ to a reciprocal $k$-nn set $N_{r, q}^k$ by only taking samples $g \in N_q^k$ into account that also contain $q$ in their own $k$-nn set $N_g^k$ (indicated by the green frames in the middle part of Figure~\ref{fig:kreciprocal}). 
Up to this step $N_{r, q}^k$ only contains samples that are already highly similar to the query image. Some gallery samples $g$ of class $c_q$ might not be directly contained in $N_{r, q}^k$, but in $N_{r, g}^k$ of some samples $g \in N_q^r,k$. Therefore, we expand $N_{r, q}^k$ to $\Tilde{N}_{r, q}^k$ by the reciprocal $\frac{1}{2}k$-nn set $N_{r, g}^{0.5 k}$ of the samples $g \in N_{r, q}^k$, if the following holds:

\begin{equation}
    |N_{r, q}^k \cap N_{r, g}^{0.5 k}| \geq \alpha |N_{r, g}^{0.5 k}|
\end{equation}

where $\alpha \in [0, 1]$, $|\Tilde{N}_{r, q}^k| = k_r$ and $k_r$ is a constant. In the expansion step of Figure~\ref{fig:kreciprocal}, samples that fulfill the above mentioned constraint for $\alpha = \frac{2}{3}$ are visualized by a green frame, those who do not by a red frame. 
If $|\Tilde{N}_{r, q}^k| < k_r$, we add the closest samples to $q$ that are not yet contained in $\Tilde{N}_{r, q}^k$. Finally, we feed $\Tilde{N}_{r, q}^k$ into the Message Passing Network to refine the feature vector of $q$. As we have shown (see Tab. 3 in the main paper) this approach improved the performance during test time by $0.5pp$ Recall@1 and $0.5pp$ NMI on CUB-200-2011 dataset and $0.5pp$ (Recall@1) and $1.4pp$ (NMI) on the Cars196 dataset.

\subsection{Teacher Student Approach}
As the latter approach requires the usage of additional parameters during test time, we develop several teacher-student approaches to transfer the knowledge of the MPN, acting as a teacher, to the backbone CNN, acting as a student. 

\noindent{\bf Knowledge Distillation.} As our approach is based on the cross-entropy loss function, we first use knowledge distillation \cite{DBLP:journals/corr/HintonVD15}, where the class probability distributions of the teacher are imitated by the student. The advantage of this technique is that these probability distributions also contain information about the similarities between classes. However, the usage of this approach does not increase the performance compared to solely using the backbone CNN, but decreases it by $4.6pp$ Recall@$1$ on CUB-200-2011 and $2.5pp$ Recall@$1$ on Cars196 (see Table~\ref{tab:settings})

\noindent{\bf Feature Imitation.} Since we are not directly interested in the class prediction quality of our network, but in the feature vectors themselves, our second approach forces the student to directly imitate the feature vectors. Further, those feature vectors of the training data after the MPN are highly discriminative. We apply the Huber Loss which is less sensitive to outliers. Again, the performance drops by $5pp$ Recall@$1$ on CUB-200-2011 and $2.5pp$ Recall@$1$ on Cars196 compared to solely using the backbone CNN.

\begin{figure}[ht]
    \begin{center}
    \centerline{\includegraphics[width=1\linewidth]{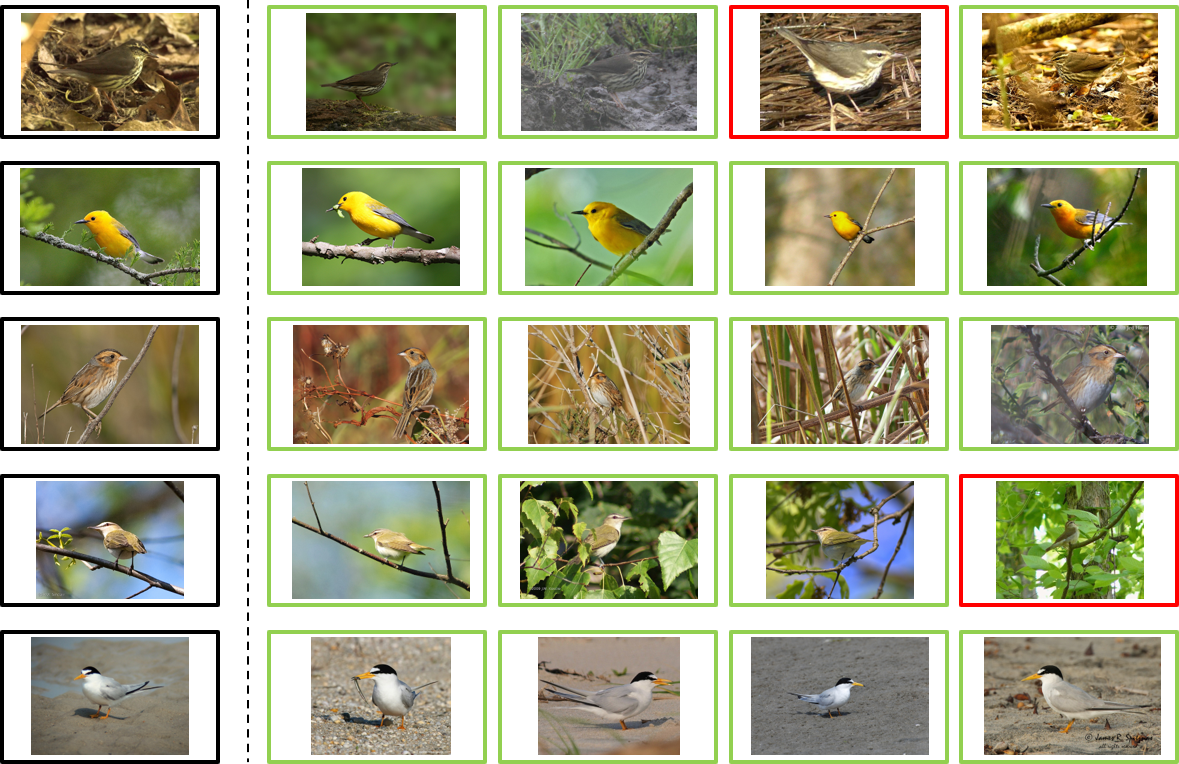}}
    \caption{More qualitative results on CUB-200-2011}
    \label{fig:more_cub}
    \end{center}
\end{figure}

\begin{figure}[ht]
    \begin{center}
    \centerline{\includegraphics[width=1\linewidth]{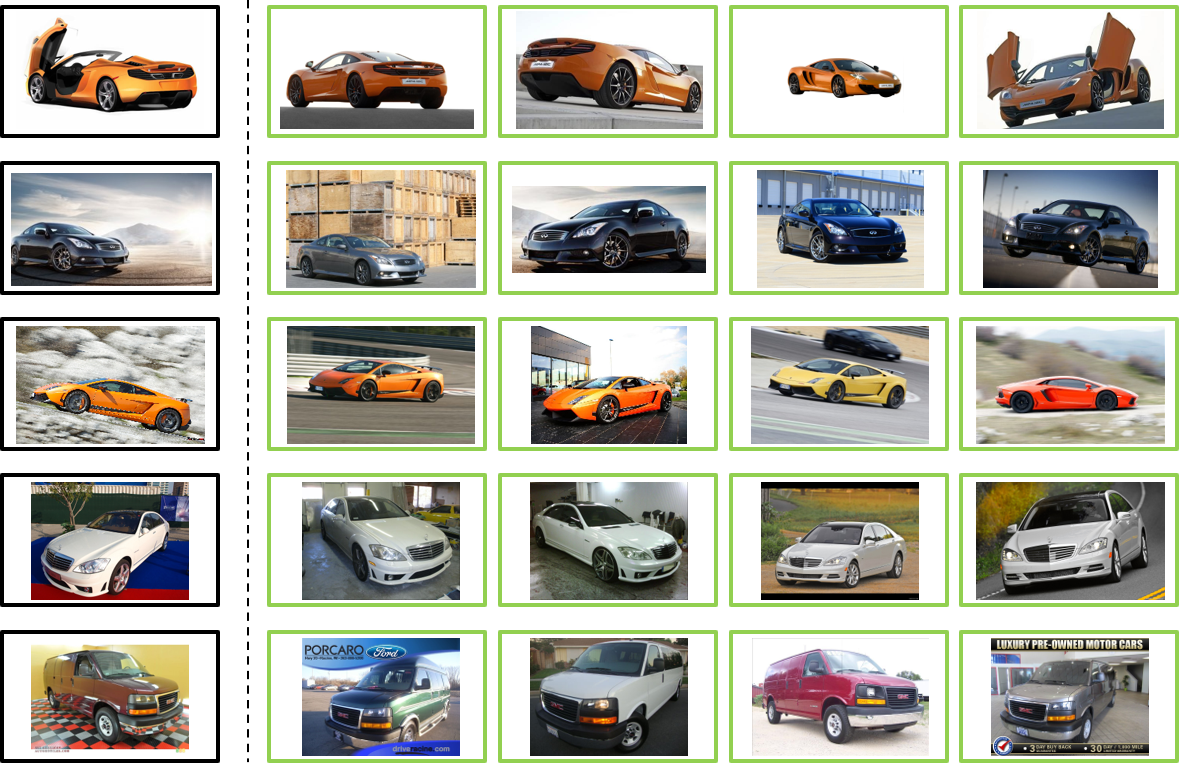}}
    \caption{More qualitative results on Cars196}
    \label{fig:more_cars}
    \end{center}
    \vspace{-0.5cm}
\end{figure}

\begin{figure}[ht]
    \begin{center}
    \centerline{\includegraphics[width=1\linewidth]{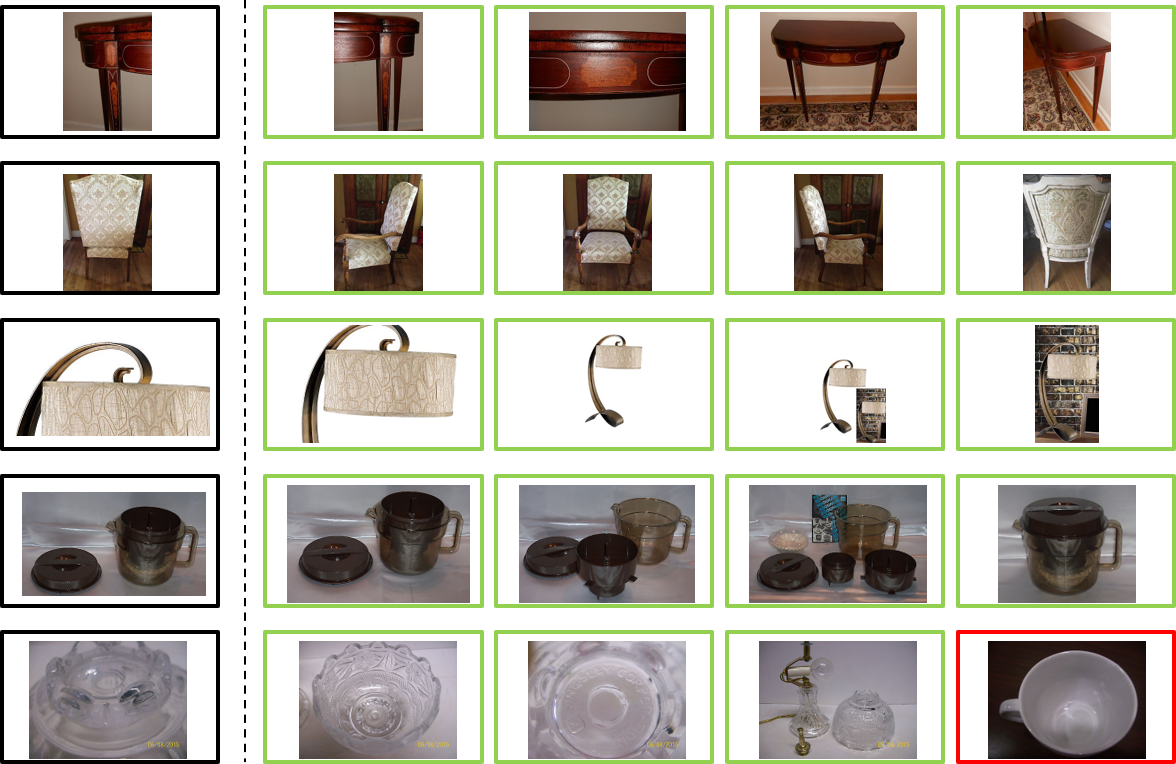}}
    \caption{More qualitative results on Stanford Online Products}
    \label{fig:more_sop}
    \end{center}
\end{figure}
\begin{figure}[ht]
    \begin{center}
    \centerline{\includegraphics[width=1\linewidth]{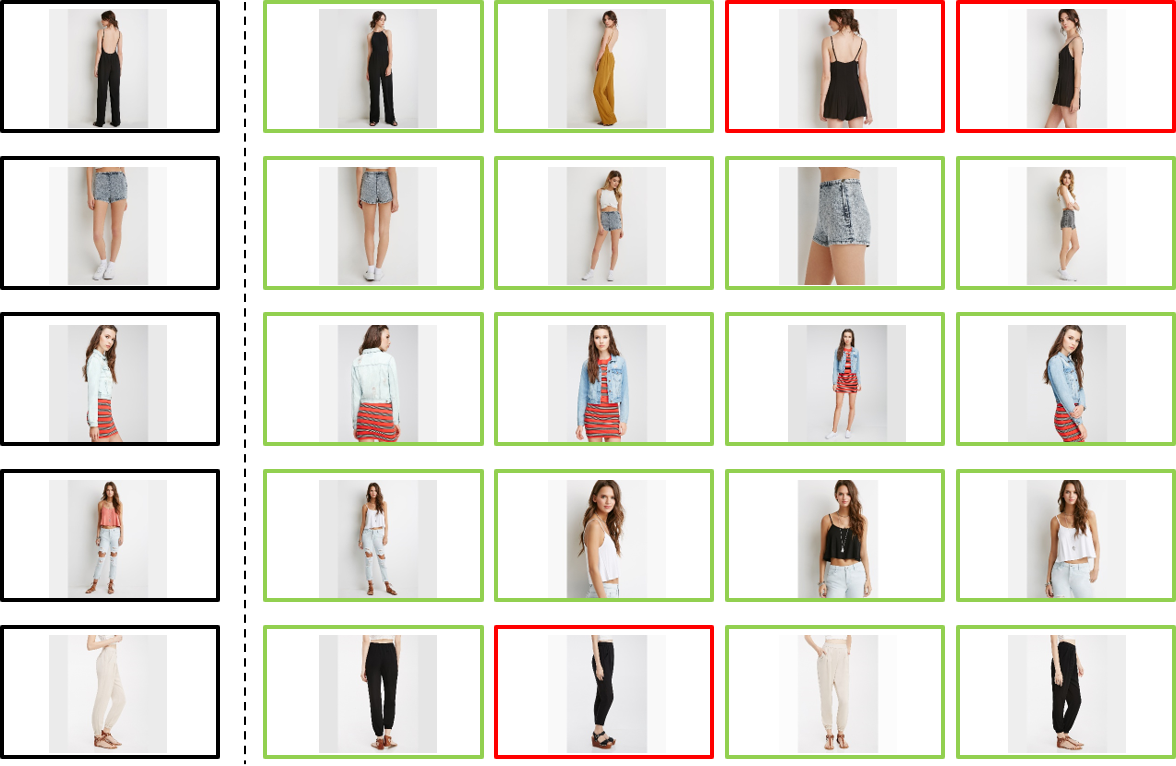}}
    \caption{More qualitative results on In-Shop}
    \label{fig:more_inshop}
    \end{center}
    \vspace{-0.5cm}
\end{figure} 
 
\noindent{\bf Relational Teacher Student.} Lastly, we utilize relational knowledge distillation \cite{DBLP:conf/cvpr/ParkKLC19}, where the student does not directly imitate the feature vectors, but the relations between different feature vectors. This also supports the paradigm of our MPN-based approach, where we refine feature vectors by taking the relations between samples into account. The performance, though, does not increase but drops even more by $5.4pp$ Recall@$1$ on CUB-200-2011 and $3.4pp$ Recall@$1$ on Cars196.

\section{Qualitative Results}
\label{sec:quali}
In the main work, we already showed several examples for our qualitative results. To show the robustness of our approach, we will now show several more samples of qualitative results on CUB-200-2011 (Figure~\ref{fig:more_cub}), Cars196 (Figure~\ref{fig:more_cars}), Stanford Online Products (Figure~\ref{fig:more_sop}) and In-Shop Clothes (Figure~\ref{fig:more_inshop}). As can be seen, our approach is able to retrieve images of the same class even for harder examples, like in the first example of CUB-200-2011 (Figure~\ref{fig:more_cub}). 

\begin{figure*}
    \centering
    \includegraphics[width=0.8\textwidth]{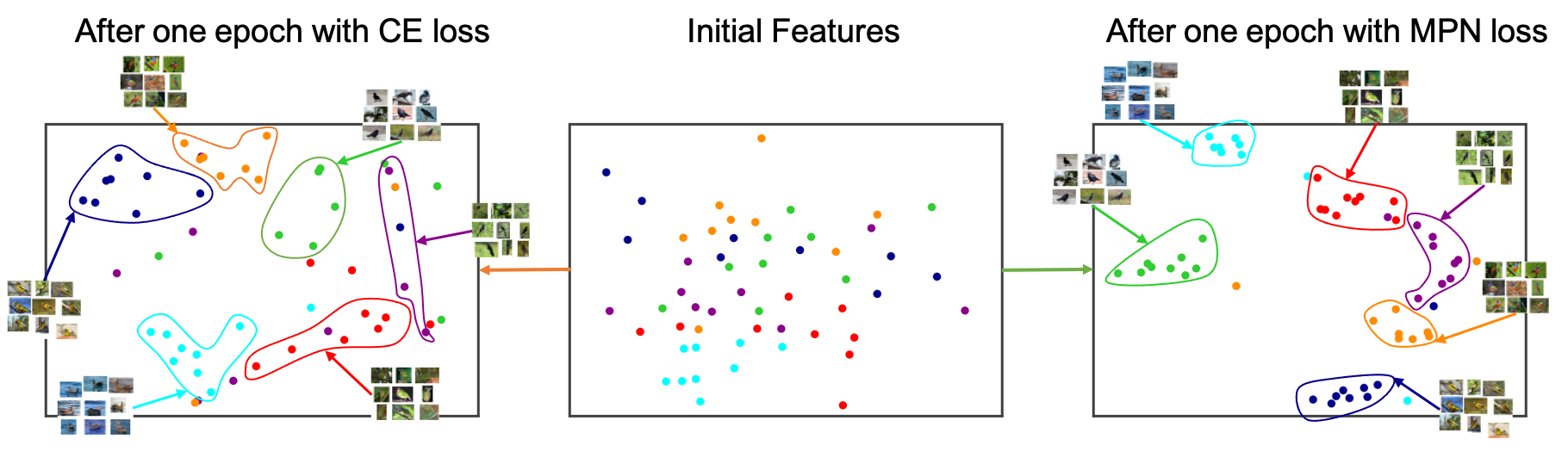}
    \caption{Comparison of the embeddings of a given batch after one epoch of training without and with MPN.}
    \label{fig:epoch_comp}
\end{figure*}

\section{T-SNE}
\label{sec:tsne}
We now show a visualization of the embedding space obtained by our approach on \textit{CUB-200-2011} using the t-distributed stochastic neighbor embedding (t-SNE) \cite{DBLP:journals/ml/MaatenH12} in Figure~\ref{fig:tsne}. Every scatter point represents a sample and different colors represent different classes. As can be seen, our approach is able to achieve representative clusters of many classes. We highlight several groups of samples, that can be best viewed when zoomed in.

\section{MPN Matters 2.0 - Comparison of Performance of Training with and without MPN}
\label{sec:MPN_matters}
In addition to the ablation studies in the main paper where we showed the performance increase of the backbone CNN trained with MPN over not using the MPN but solely cross-entropy loss during training (see MPN matters) as well as some visualizations of how the class prediction after the MPN is influenced by other samples in the batch, we provide some more visualizations that support these findings. 

Firstly, we visualize the difference between the embeddings after the backbone CNN of a given batch after one epoch of training without and with MPN using t-distributed stochastic neighbor embedding (t-SNE) \cite{DBLP:journals/ml/MaatenH12}. As can be seen in Figure~\ref{fig:epoch_comp}, the features after the backbone CNN are much more clustered than when training without MPN.

\begin{figure}[ht!]
    \begin{center}
    \centerline{\includegraphics[width=1\linewidth]{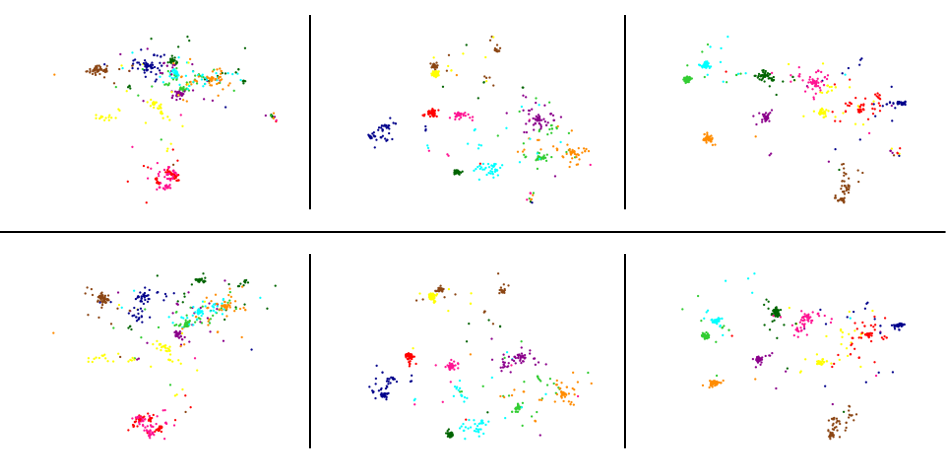}}
    \caption{Visualization of 10 sampled classes from CUB-200-2011 test dataset when trained without MPN (upper row) and with MPN (lower row).}
    \label{fig:cub_vis}
    \end{center}
\end{figure}

\begin{figure}[ht!]
    \begin{center}
    \centerline{\includegraphics[width=1\linewidth]{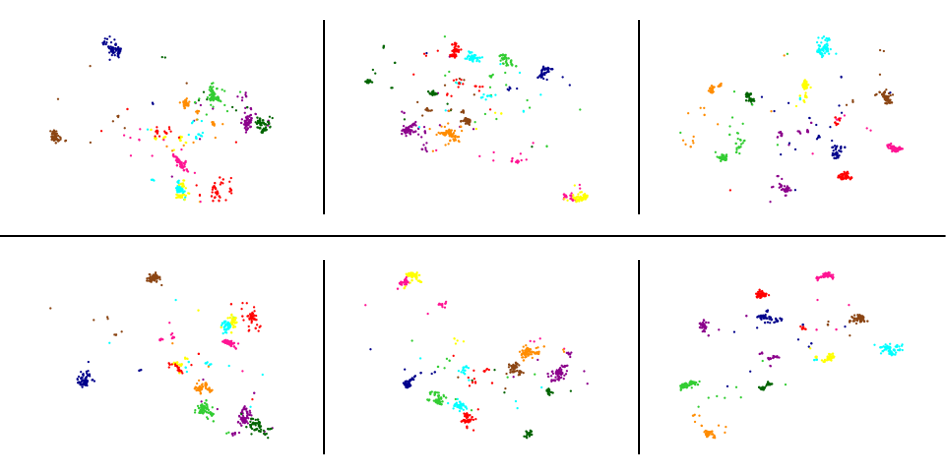}}
    \caption{Visualization of 10 sampled classes from Cars196 test dataset when trained without MPN (upper row) and with MPN (lower row).}
    \label{fig:cars_vis}
    \end{center}
    \vspace{-0.7cm}
\end{figure}

Secondly, we again utilize t-SNE \cite{DBLP:journals/ml/MaatenH12} to get low dimensional representations of the embeddings of all samples in a given test set after the whole training without as well as with MPN and sample 10 classes for the sake of clarity. In Figure~\ref{fig:cub_vis} and Figure~\ref{fig:cars_vis} we show three such subsets of classes for CUB-200-2011 and Cars196, respectively. The upper rows in both figures represent embeddings generated by the backbone CNN trained without MPN while the lower ones show embeddings generated by the backbone trained with MPN. 
The backbone CNN trained using solely cross-entropy loss performs well on many samples. However, our approach is able to better divide more difficult classes from the remaining classes in the embedding space as can be seen for example from the dark blue class in the first column of Figure~\ref{fig:cub_vis}. 
Further, it is less prone to outliers as can be seen in the second column in Figure~\ref{fig:cars_vis}, where there are fewer outliers in all classes in the lower row that shows the visualizations of the embeddings trained with MPN. 
Finally, the embeddings of samples of the same class most often lie closer together and are further apart from other classes as can be seen from the red and pink classes in the central column of Figure~\ref{fig:cub_vis} or the dark blue, orange, and pink classes in the first column of Figure~\ref{fig:cars_vis}.

\section{Analysis of Number of Message Passing Steps and Heads on Stanford Online Products}
\label{sec:LH_SOP}
\begin{figure}
  \begin{center}
  \centerline{\includegraphics[width=1\linewidth]{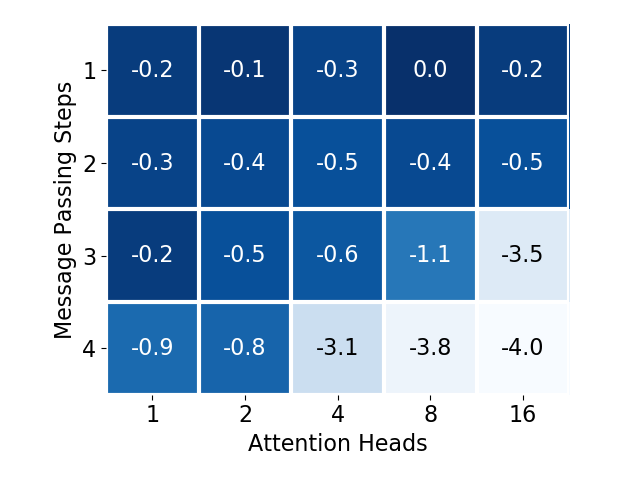}}
  \caption{Relative Difference with respect to Best Recall@1 on Stanford Online Products.} \label{fig:sop_LH}
  \end{center}
  \vspace{-0.3cm}
\end{figure}

Additionally to the analysis of different numbers of message passing steps and attention heads on CUB-200-2011 and Cars196 in the main paper, we provide an equal analysis on Stanford Online Products. Investigating the results on CUB-200-2011 and Cars196 datasets one could assume that with increasing size of the dataset an increasing number of message passing steps is needed, as the best performing model on CUB-200-2011 utilizes only one message passing step while the best performing model on Cars196 utilizes two message passing steps and CUB-200-2011 is smaller than Cars196. However, as can be seen in Figure~\ref{fig:sop_LH} this is not the case, and the performance of our approach on Stanford Online Products drops with an increasing number of message passing steps and attention heads. As already mentioned in the main paper, this is in line with \cite{DBLP:conf/iclr/VelickovicCCRLB18}, who also utilize few message passing steps when applying graph attention. 

\begin{figure*}[ht!]
\centering
\begin{minipage}{.48\textwidth}
  \begin{center}
  \centerline{\includegraphics[width=1\linewidth]{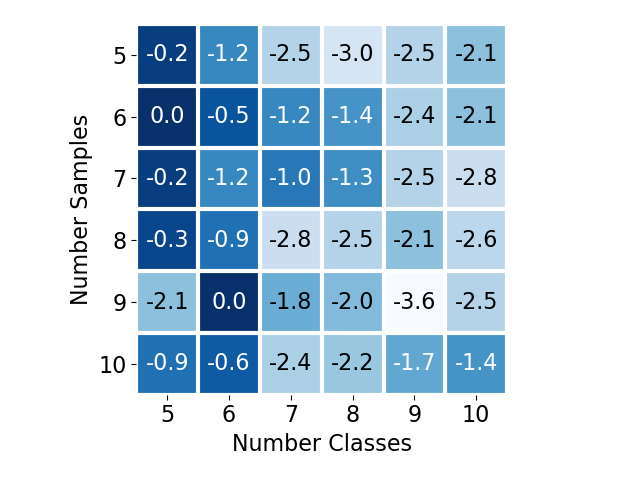}}
  \caption{Relative Difference with respect to Best Recall@1 on CUB-200-2011.} \label{fig:cub_CS}
  \end{center}
\end{minipage}%
\hfill
\begin{minipage}{.48\textwidth}
  \begin{center}
  \centerline{\includegraphics[width=1\linewidth]{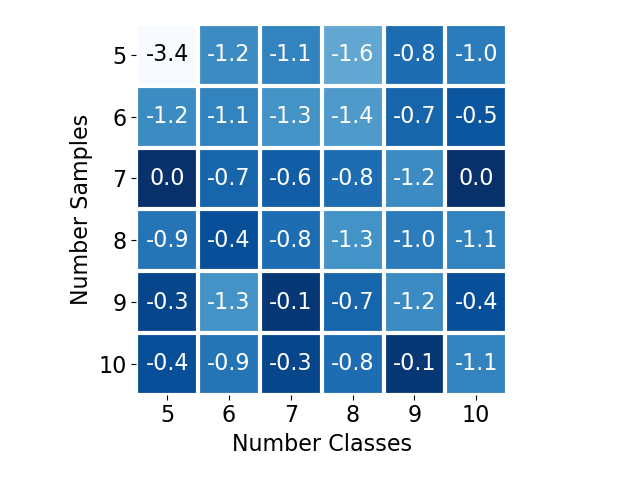}}
  \caption{Relative Difference with respect to Best Recall@1 on Cars196.} \label{fig:cars_CS}
  \end{center}
\end{minipage}%
\end{figure*}
\section{Different Number of Classes}
\label{sec:CS}
We also conduct experiments to investigate the impact of the composition of the batches concerning the number of classes and the number of samples per class. 
Therefore we vary the number of classes and samples on CUB-200-2011 and Cars196 between five and ten. As can be seen in Figure~\ref{fig:cub_CS}, the performance on CUB-200-2011 tends to go down with an increasing number of classes while on Cars196 (see Figure~\ref{fig:cars_CS}) the performance drops when only a few samples per class are used. However, it can be said that the performance is stable with the biggest drop in performance on CUB-200-2011 being $2.8pp$ and $3.4pp$ on Cars196.

\vspace{-0.1cm}
\section{Metric Learning Reality Check}
\label{sec:Reality_Check}
\cite{DBLP:conf/eccv/MusgraveBL20} claim that the huge improvements of recent metric learning approaches over prior works is mainly caused by flaws in the experimental methodology like utilizing a more powerful backbone, unsuitable evaluation metrics or training with test set feedback. The authors show, that the performance of ResNet50 is worse on CUB-200-2011 and Cars196 than when using BN-Inception. To prove that our approach is robust to the hyperparameter choice we followed \cite{DBLP:conf/eccv/MusgraveBL20}  to find the hyperparameters (e.g., number of epochs) without feedback from the test set and report the results here. We get state-of-the-art results in Cars196, Stanford Online Products, and In-Shop datasets, and competitive results on CUB-200-2011 dataset compared to results mentioned in \cite{DBLP:conf/eccv/MusgraveBL20} as can be seen in Table~\ref{tab:reality_check}.

\input{tables/reality_check}
\vspace{-0.2cm}
\section{Large Number of Classes}
\label{sec:large_num_classes}

Our method uses a fully connected layer to compute the final loss function. In cases where the number of classes increases, then the size of the classification layer increases too. While this typically is not a problem for metric learning datasets which contain from hundreds to tens of thousands of classes, it can become a problem for the closely related problem of face recognition, where datasets typically contain millions of classes. Unlike our method (and other classification-based methods \cite{DBLP:journals/corr/abs-1811-12649, DBLP:conf/aaai/ZhengJSZWH19, DBLP:journals/corr/abs-1909-05235, DBLP:conf/eccv/GrLoss}), methods that use a pairwise (e.g.\, contrastive/triplet) loss function do not have this problem.

Nevertheless, there are ways of facing the problem. Normalized Softmax \cite{DBLP:journals/corr/abs-1811-12649} tackles the problem by sampling a mini-batch only from a certain number of classes, a strategy proposed also in Group Loss \cite{DBLP:conf/eccv/GrLoss}. Our method uses this sampling strategy in default mode (in each mini-batch, we sample only a certain number of classes). Consequently, we know in advance which units of the last layer need to be modified, and all the other units can easily get masked out for a more efficient tensor-tensor multiplication.

Dealing with datasets that contain a large number of classes is a problem that has been widely studied in natural language processing \cite{DBLP:journals/corr/abs-1301-3781}, typically solved by replacing the softmax layer with hierarchical-softmax \cite{DBLP:conf/nips/MnihH08}. Considering that the problem is similar, we could envision replacing softmax with hierarchical-softmax for our problem to have a more efficient method.

\begin{figure*}
    \begin{center}
    \centerline{\includegraphics[width=0.99\textwidth]{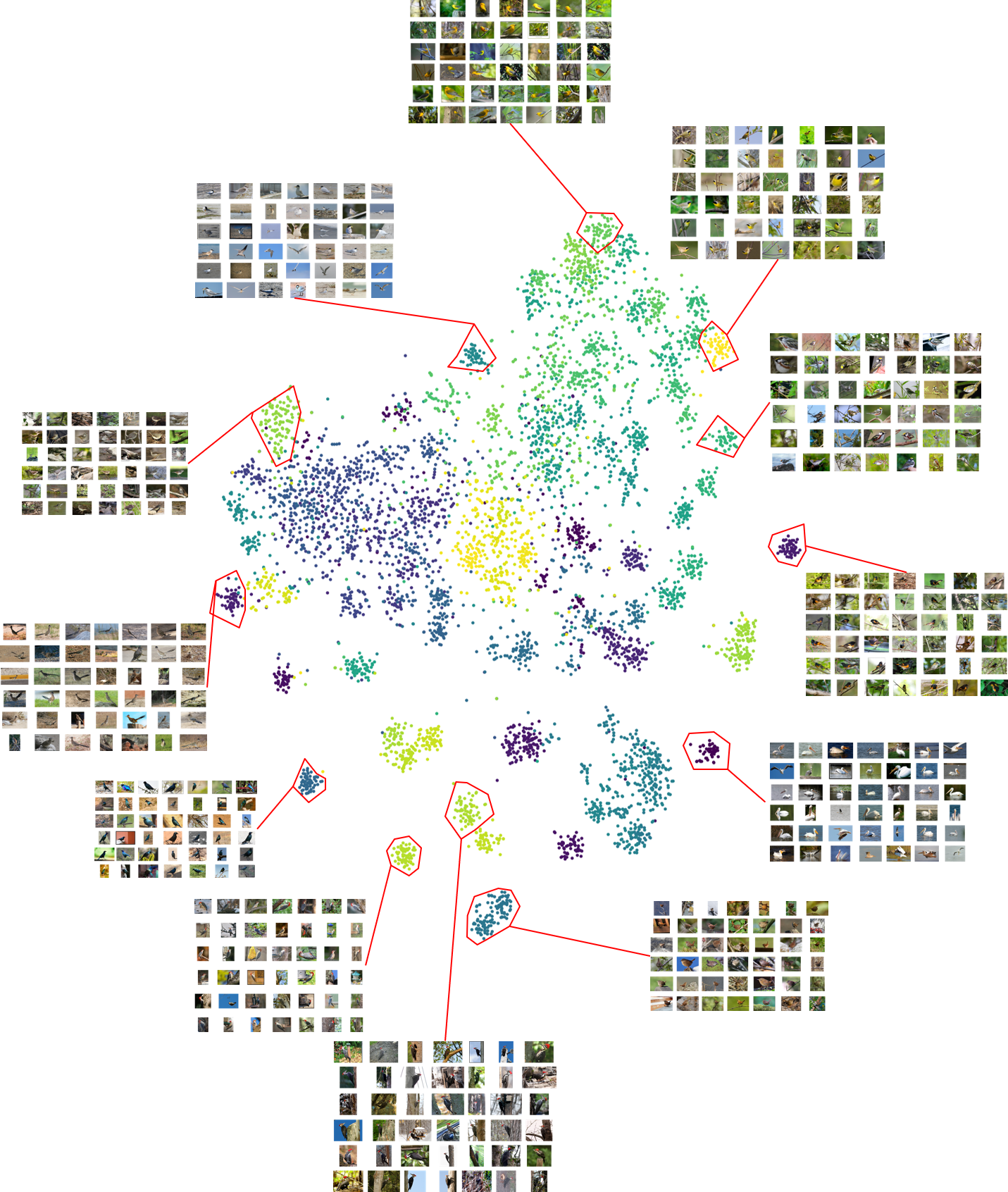}}
    \caption{ t-SNE \cite{DBLP:journals/ml/MaatenH12} visualization of our  embeddings on the CUB-200-2011 \cite{WahCUB_200_2011} dataset with some clusters highlighted. Best viewed on a monitor when zoomed in.}
    \label{fig:tsne}
    \end{center}
\end{figure*}

\clearpage

%% file: tables/embedding_size.tex
\begin{table*}[hbt!]
\centering
\resizebox{0.9\textwidth}{!}{
\begin{tabular}{@{}l|cc|cc|cc|cc}
\hline
& \multicolumn{2}{c}{CUB-200-2011} & \multicolumn{2}{c}{Cars196} & \multicolumn{2}{c}{Stanford Online Products} & \multicolumn{1}{c}{In-Shop Clothes} \\ 
\hline
 & \textbf{R@1} &  \textbf{NMI} & \textbf{R@1} &  \textbf{NMI} & \textbf{R@1} &  \textbf{NMI} & \textbf{R@1}\\
\hline
Dimension 64 & 61.8 & 68.8 & 76.3 & 68.1 & 73.8 & 91.2 & 87.5 \\
Dimension 128 & 65.1 & 69.3 & 81.6 & 69.4 & 79.1 & 92.3 & 91.3 \\
Dimension 256 & 68.4 & 73.4 & 86.3 & 72.1 & 80.8 & 92.6 & 92.5 & \\
Dimension 512 & 70.3 & \textbf{74.0} & 88.1 & 74.8 & \textbf{81.4} & \textbf{92.6} & \textbf{92.8} \\
Dimension 1024 & \textbf{71.8} & 72.8 & \textbf{89.5} & \textbf{75.0} & \textbf{81.4} & 92.5 & 91.7 \\
\hline
\end{tabular}}
\caption{Performance of our approach using different embedding sizes on CUB-200-2011, Cars196, Stanford Online Products and In-Shop Clothes datasets.}
\label{tab:embedding_dim}
\end{table*}


%% file: tables/proxynca++.tex
\begin{table*}[htp]
    \centering
    \resizebox{0.9\textwidth}{!}{
    \begin{tabular}{@{}l|c|c|c|c@{}}
        \hline
         & \multicolumn{1}{c}{CUB-200-2011} & \multicolumn{1}{c}{Cars196} & \multicolumn{1}{c}{Stanford Online Products} & \multicolumn{1}{c}{In-Shop Clothes} \\
         \hline
         Our results on \cite{DBLP:journals/corr/abs-2004-01113} & 66.3 & 84.9 & 79.8 & 90.4 \\
         Results in \cite{DBLP:journals/corr/abs-2004-01113} & 64.7 $\pm$ 1.6 & 85.1 $\pm$ 0.3 & 79.6 $\pm$ 0.6 & 87.6 $\pm$ 1.0 \\
         \hline
    \end{tabular}}
    \caption{Comparison of Recall@$1$ on images of size $227 \times 227$ using ProxyNCA ++ \cite{DBLP:journals/corr/abs-2004-01113} of the results reported in \cite{DBLP:journals/corr/abs-2004-01113} and our results obtained by running their code.}
    \label{tab:proxynca++}
\end{table*}

%% file: tables/larger_images.tex
\begin{table*}[hbt!]
\centering
\resizebox{0.9\textwidth}{!}{
\begin{tabular}{@{}l|cc|cc|cc|cc@{}}
\hline
& \multicolumn{2}{c}{CUB-200-2011} & \multicolumn{2}{c}{Cars196} & \multicolumn{2}{c}{Stanford Online Products} & \multicolumn{1}{c}{In-Shop Clothes} \\
\hline
 & \textbf{R@1} &  \textbf{NMI} & \textbf{R@1} &  \textbf{NMI} & \textbf{R@1} &  \textbf{NMI}& \textbf{R@1}\\
\hline
Horde$^{512\dagger}$ \cite{DBLP:journals/corr/abs-1908-02735} & 66.3 & - & 83.9 & - & 80.1 & - & 90.4 \\
Proxy NCA++$^{512\dagger}$ \cite{DBLP:journals/corr/abs-2004-01113} & 69.0 & 73.9 & 86.5 & 73.8 & 80.7 & - & 90.4\\
Proxy Anchor$^{512\dagger}$ \cite{DBLP:conf/cvpr/KimKCK20} & 71.1 & - & 88.3 & - & 80.3 & - & 92.6 \\ \hline
Ours$^{512}$ & 70.3 & 74.0 & 88.1 & 74.8 & 81.4 & 92.6 & 92.8 \\
Ours$^{512\dagger}$ & \textbf{71.7} & \textbf{74.3} & \textbf{90.2} & \textbf{75.4} & \textbf{81.7} & \textbf{92.3} & \textbf{92.9}\\
\hline
\end{tabular}}
\caption{Performance of our approach using larger images compared to the approaches that only report their results on larger images. $\dagger$ indicates results on larger images. Proxy Anchor \cite{DBLP:conf/cvpr/KimKCK20} presents the results both in regular (shown in the tables in the main paper) and large size images.}
\label{tab:larger_img}
\end{table*}

%% file: tables/eval_settings.tex
\begin{table}[hbt!]
\centering
\resizebox{0.45\textwidth}{!}{
\begin{tabular}{@{}l|cc|cc@{}}
\hline
& \multicolumn{2}{c}{CUB-200-2011} & \multicolumn{2}{c}{CARS196} \\ 
\hline
 & \textbf{R@1} &  \textbf{NMI} & \textbf{R@1} &  \textbf{NMI}\\
\hline
backbone only & 70.3 & 74.0 & 88.1 & 74.8 \\ 
\hline
K-means$^\dagger$ & 66.1 & 69.6 & 85.1 & 70.5 \\
Ward clustering$^\dagger$ & 67.9 & 68.8 & 86.6 & 69.5 \\
Spectral Clustering$^\dagger$ & 63.1 & 69.6 & 85.1 & 69.2  \\
Birch$^\dagger$ & 65.5 & 69.5 & 86.1 & 70.0 \\
DBSCAN* & 65.7 & 72.0 & 85.0 & 70.3 \\
Optics* & 66.5 & 71.1 & 85.4 & 69.6 \\
\hline
Nearest neighbors & 62.1 & 69.2 & 84.8 & 70.2 \\ \hline
Reciprocal kNN & \textbf{70.8} & \textbf{74.5} & \textbf{88.6} & \textbf{76.2} \\
\hline 
Knowledge Distillation & 65.7 & 70.3 & 85.6 & 71.6 \\
Feature Imitation & 65.3 & 70.1 & 85.6 & 70.9 \\
Relational TS & 64.9 & 68.7 & 84.7 & 70.5 \\
\hline
\end{tabular}}
\caption{Performance of different settings of using the MPN during test time as well as the performance of teacher student approaches. $\dagger$ indicates clustering algorithms that need a fixed number of clusters (900 clusters), * indicates density-based clustering algorithms (eps=0.9, min sample=5)}
\label{tab:settings}
\vspace{0.3cm}
\end{table}

%% file: tables/reality_check.tex
\begin{table}
\centering
\resizebox{0.49\textwidth}{!}{
\begin{tabular}{l|c|c|c|l}
\hline
 & CUB & CARS & SOP & In-Shop \\ \hline
Ours&  \textbf{70.3}   &  \textbf{88.1}    &  \textbf{81.4}   &    \textbf{92.8}    \\  \hline
Ours reality check & 67.1$\pm$0.69 & 86.7$\pm$0.48 & 81.1$\pm$0.13 & 92.5$\pm$0.11 \\\hline
\end{tabular}}
\caption{Performance of our approach following \cite{DBLP:conf/eccv/MusgraveBL20} to find the hyperparameters (e.g., number of epochs) without feedback from the test set.}
\label{tab:reality_check}
\end{table}